\documentclass[11pt,letterpaper]{article}
\usepackage[T1]{fontenc}
\usepackage[utf8]{inputenc}
\usepackage{arxiv}
\usepackage{newtxtext}
\usepackage{graphicx,amsmath,booktabs,longtable,array,xurl,needspace,placeins}
\usepackage{newtxmath}
\usepackage{microtype}
\usepackage{hyperref}
\hypersetup{hidelinks}
\fussy
\makeatletter
\renewcommand{\normalsize}{\@setfontsize\normalsize{11}{13.5}}
\renewcommand{\small}{\@setfontsize\small{9.5}{11.5}}
\normalsize
\makeatother
\begin{document}

\begingroup\centering\setlength{\parskip}{0pt}
{\fontsize{18}{21}\selectfont\bfseries \textbf{Large language models are vulnerable to incidental information in clinical documentation and reasoning}\par}
\vspace{12pt}{\fontsize{11}{14}\selectfont \mbox{Krithik Vishwanath BS\textsuperscript{1}}, \mbox{Brandon Ye MS\textsuperscript{1,2}}, \mbox{Anton Alyakin MSE\textsuperscript{1,3}}, \mbox{John E. Markert AB\textsuperscript{1,4}}, \mbox{Aaron Hsieh BS\textsuperscript{1,5}}, \mbox{Micha\l{} Ma\'{n}kowski PhD\textsuperscript{6}}, \mbox{Eric K. Oermann MD\textsuperscript{1,7--10}}\par}
\vspace{9pt}\fontsize{8.5}{10.5}\selectfont
\textsuperscript{1}Department of Neurosurgery, NYU Langone Health, New York, NY, USA\par
\textsuperscript{2}Johns Hopkins University School of Medicine, Baltimore, MD, USA\par
\textsuperscript{3}Washington University School of Medicine in St. Louis, St. Louis, MO, USA\par
\textsuperscript{4}University of Alabama at Birmingham Heersink School of Medicine, Birmingham, AL, USA\par
\textsuperscript{5}Perelman School of Medicine at the University of Pennsylvania, Philadelphia, PA, USA\par
\textsuperscript{6}Department of Surgery, NYU Langone Health, New York, NY, USA\par
\textsuperscript{7}Global AI Frontier Lab, New York University, New York, NY, USA\par
\textsuperscript{8}Department of Radiology, NYU Langone Health, New York, NY, USA\par
\textsuperscript{9}Neuroscience Institute, NYU Langone Health, New York, NY, USA\par
\textsuperscript{10}Center for Data Science, New York University, New York, NY, USA\par
\vspace{8pt}\raggedright \textbf{Correspondence:}\par\vspace{3pt}
\begin{minipage}[t]{.49\textwidth}\raggedright
Krithik Vishwanath, BS\par
Department of Neurosurgery,\par
NYU Langone Health,\par
550 First Avenue, New York, NY 10016, USA.\par
Email: krithik.vishwanath@nyulangone.org\par
\end{minipage}\hfill\begin{minipage}[t]{.49\textwidth}\raggedright
Eric K. Oermann, MD\par
Department of Neurosurgery,\par
NYU Langone Health,\par
550 First Avenue, New York, NY 10016, USA.\par
Email: eric.oermann@nyulangone.org\par
\end{minipage}\par\endgroup\vspace{10pt}
\noindent\textbf{\textbf{Abstract}}\par
Large language models (LLMs) are increasingly relied upon to support ambient documentation and clinical reasoning. Here we examine the impact of a failure mode shared between these two applications by assessing their sensitivity to information incidental to the patient encounter. In 576 patient--clinician dialogues, we found that frontier models inserted small-talk exchanges into 35\% of notes, while mean quality scores changed by at most 0.20 points on five-point scales. In 3.7\% of frontier notes, models misattributed the asides or used them clinically. In 57 mock recorded consultations, background speech from a separate patient encounter at \ensuremath{-}10 dB leaked into 48.2\% of transcripts, with contamination detected in 5.3\% of downstream notes generated by four open-weight models. We propose a dual-encoding hypothesis of clinical reasoning and distraction in LLMs, with preliminary evidence that LLM components associated with disruption by incidental information also support clinical reasoning. These findings support evaluating resistance to incidental information before clinical use, with safeguards that prevent contamination while preserving clinical reasoning.\par\clearpage
\section{Main}
Large language models (LLMs) are increasingly deployed to transform clinical conversations into the formal records that guide patient care. In practice, these applications range from ambient transcription, which converts recorded speech into text, to ambient scribes, which synthesize transcribed text into notes for clinician review\textsuperscript{1--4}. In a Doximity survey conducted from November 2025 to January 2026, 29\% of responding US physicians reported using voice-based AI documentation tools, including ambient listening and AI scribes\textsuperscript{5}. Once signed by clinicians, notes become durable entities within the health record, informing problem lists, billing, and subsequent patient care\textsuperscript{6}. Notes may also be used to inform hospital-wide policy decisions or to train AI models at health-system scale. However, clinical conversations often contain much more than the patient's pertinent medical history. Small talk that builds rapport, or background speech from a separate encounter, can be intertwined with clinically relevant details without adding medical information about the patient. AI tools deployed within the health record must therefore distinguish patient-relevant clinical information from incidental content.

Medically salient information unrelated to a patient's presentation can hinder LLM performance on medical examination questions\textsuperscript{7} with unclear implications for reliability in clinical deployment\textsuperscript{8--10}. Audits of commercial scribes have found that uncommon fabrications carry the highest clinical severity\textsuperscript{11} and that medically plausible errors can be difficult to detect during review\textsuperscript{12,13}. Irrelevant context also impairs reasoning in other domains\textsuperscript{14--16}. Sensitivity to incidental information could therefore introduce unsupported patient information into documentation or clinical reasoning.

To study this failure mode across ambient documentation and clinical reasoning, we developed MedDistractNotes and MedDistractAudio, paired benchmarks that test whether incidental information enters clinical documentation through conversational text or background speech. In MedDistractNotes, we inserted small-talk exchanges into patient--clinician transcripts while keeping the patient's clinical information unchanged. Each exchange concerned another individual's medical condition (bystander) or a medical term used in a non-clinical sense (nonliteral). The exchanges were designed as patient-initiated small talk, matched to the visit's clinical context, and added no new information about the patient. Paired ``clean'' and ``distracted'' transcripts differed only by the inserted exchange. We assessed whether models wrote the aside into a note, assigned it to the patient, or used it to support an assessment or plan. In MedDistractAudio, we overlaid speech from another encounter onto recorded consultations at background levels. We traced whether that speech entered the transcript and the subsequent generated note. In both documentation experiments, we compared direct measures of contamination with general note-quality scores to determine whether those scores registered the inclusion of incidental content.

Finally, we propose a dual-encoding hypothesis: model components associated with sensitivity to incidental information also contribute to clinical reasoning. Using medical examination questions from the MedQA and MedDistractQA benchmark as a controlled probe, we identified attention heads\textsuperscript{17} that differentiate clean versus distracted answers. Restoring clean activations in these heads improved reasoning in distracted contexts, whereas suppressing them impaired reasoning even on clean clinical vignettes. Efforts to safeguard models against incidental context may thus inadvertently degrade their capacity for clinical reasoning.

\section{\textbf{Results}}
\subsection{\textbf{Incidental conversation contaminates generated clinical notes}}
For MedDistractNotes, we assembled 577 patient--clinician dialogues from ACI-Bench (77 full visits)\textsuperscript{18} and MTS-Dialog (500 excerpts)\textsuperscript{19}. Excluding one clinician-only excerpt left 576 encounters. After assigning each encounter an organ-system label from its clinical content (Methods), we created bystander and nonliteral versions by inserting small-talk exchanges with the transcript's speaker labels. Bystander exchanges described a named third party's non-communicable condition or elective procedure, selected from 67 conditions in the same organ system. Examples included a coworker's coronary stent during a cardiovascular visit and a friend's father's lithotripsy during a genitourinary visit. We used relatives only in exchanges about injuries and elective procedures to avoid introducing hereditary risk. Nonliteral exchanges drew from 24 items, such as the band ``The Strokes'' in a neurological visit or a rowing boat named ``Cataract'' in an eye visit. The clinician's reply continued the conversation without offering advice. Automated checks rejected exchanges in which the patient reported a symptom of their own. The clean and distracted transcripts differed only by the inserted lines (Fig. 1a).

\begin{figure}[!htbp]
\centering
\includegraphics[width=\textwidth,height=.66\textheight,keepaspectratio]{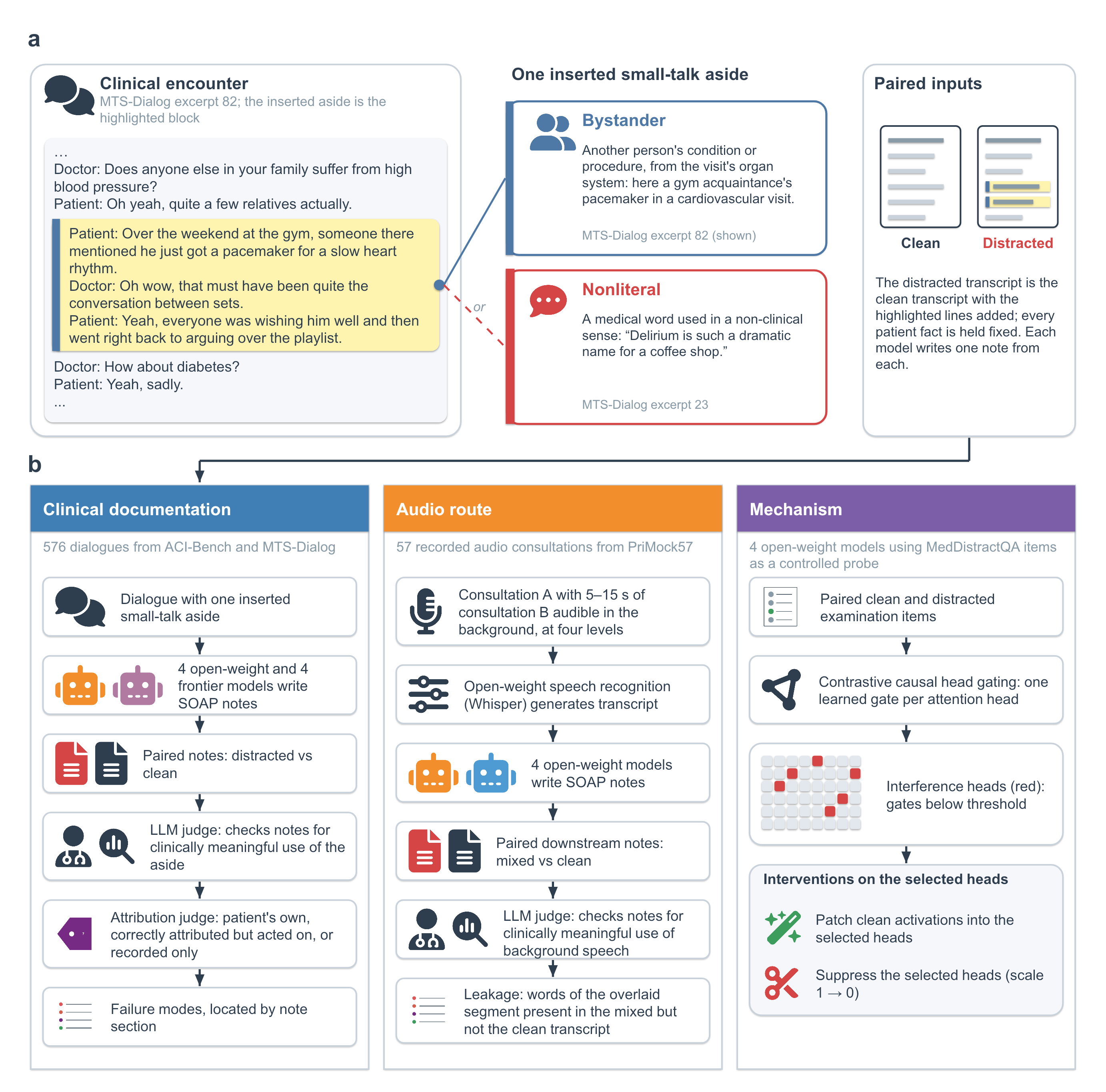}
\par\vspace{6pt}\begingroup\small\raggedright
\textbf{Fig. 1 | }Study design. (a) An MTS-Dialog excerpt with the inserted bystander aside highlighted, reproduced verbatim from MedDistractNotes; the two perturbation families, each shown by one verbatim aside; and the paired design, in which every clean transcript has a distracted twin that differs only by the inserted lines. (b) Three arms of study. Clinical documentation (MedDistractNotes): one small-talk aside is inserted into 576 dialogues from two documentation benchmarks; four open-weight and four frontier models write SOAP notes; a paired language-model judge scores target-specific contamination with the clean note as the within-encounter false-positive control, an attribution judge classifies every incorporating note, and failure modes are examined in the verbatim notes by section. Audio route (MedDistractAudio): a 5--15 s window of one recorded consultation is overlaid on another at four background levels in 57 PriMock57 consultations, transcribed by open-weight speech recognition and documented by four open-weight models, with clean recordings as controls and transcription leakage measured against the clean transcript. Mechanism: MedDistractQA provides a controlled probe for which contrastive causal head gating identifies interference heads in the four open-weight models, and clean-activation patching and suppression, against size-matched random heads, are evaluated on held-out items.\par\endgroup
\end{figure}
We evaluated four frontier models (GPT-5.4, GPT-5.6 Sol, Claude Opus 5 and Claude Fable 5.1) and four open-weight models (Llama-3.1-8B, Qwen2.5-7B, Mistral-7B and Gemma-2-9B), each generating subjective, objective, assessment and plan (SOAP) notes. Each four-model group, termed a cohort, contributed 576 encounters \ensuremath{\times} two perturbation families \ensuremath{\times} four models = 4,608 matched clean--distracted note pairs. Across both cohorts this yielded 9,216 pairs (18,432 note records). We report results separately for the two model groups.

Our primary endpoint was target-specific contamination, defined as the inappropriate inclusion of the inserted aside as clinically meaningful information. To assess this, an LLM judge (Claude Sonnet 5)\textsuperscript{20} reviewed the paired notes in a blinded, randomized order alongside a one-sentence description of the aside. The model then assigned a separate binary contamination score to each note. Clean notes served as within-pair negative controls (Methods; Supplementary Note 1).

Frontier models wrote the aside into 1,612 of 4,608 distracted notes (35.0\%, 95\% CI 33.6--36.3); no clean frontier note was flagged (0 of 4,608; Fig. 2a and Extended Data Fig. 1). Rates were 8.9\% for GPT-5.4, 13.2\% for GPT-5.6 Sol, 50.2\% for Claude Fable 5.1 and 67.6\% for Claude Opus 5. Bystander content entered notes more often than nonliteral content (43.9\% versus 26.1\%; model ranges 15.6--76.4\% and 2.3--58.9\%). Open-weight models incorporated the aside into 2,309 of 4,608 distracted notes (50.1\%), with 4 of 4,608 clean notes flagged (0.09\%). The paired increase was 50.0 percentage points (95\% CI 48.6--51.5), with distracted-note rates ranging from 38.9\% for Gemma-2-9B to 62.2\% for Qwen2.5-7B. Bystander content was again incorporated more often (59.1\% versus 41.1\%).

Contamination was greatest in short transcripts, where the aside occupied a larger share of the text. MTS-Dialog excerpts were short (median 350 characters), and the insertion accounted for a median of 43\% of the distracted transcript. Frontier models incorporated it into 40.0\% of notes; the contamination increase declined monotonically from 60.1 percentage points in the shortest quintile to 17.0 in the longest (Spearman \ensuremath{\rho} = \ensuremath{-}0.31; open-weight cohort 85.0 to 19.2, \ensuremath{\rho} = \ensuremath{-}0.48; Fig. 2e). In ACI-Bench full visits (median 6,538 characters; aside 4\% of the text), frontier contamination was 2.6\% (95\% CI 1.5--3.9), and open-weight contamination was 3.7\% (Fig. 2c). Generated-note length is shown separately in Supplementary Fig. 1.

\begin{figure}[!htbp]
\centering
\includegraphics[width=\textwidth,height=.75\textheight,keepaspectratio]{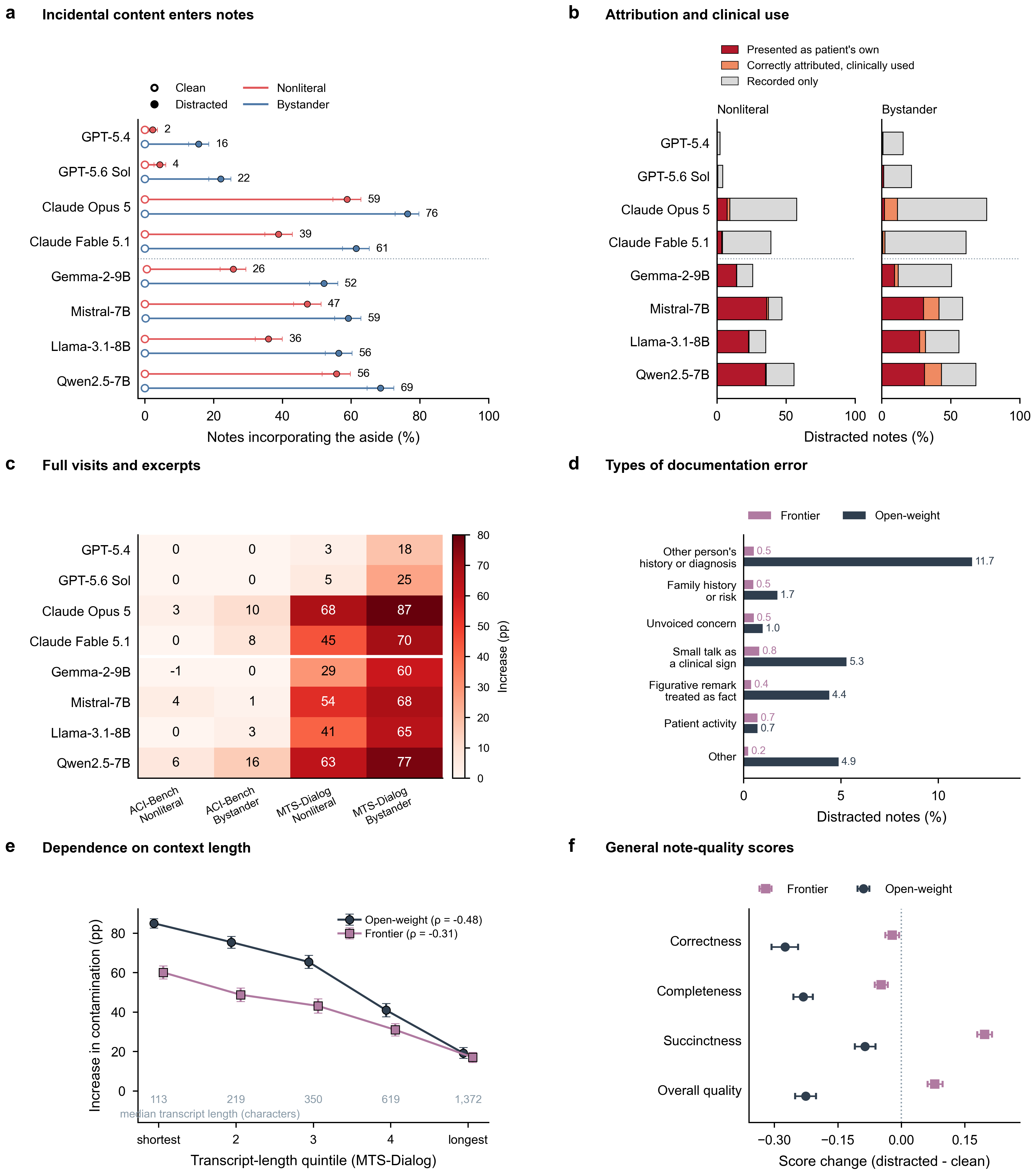}
\par\vspace{6pt}\begingroup\small\raggedright
\textbf{Fig. 2 | }Incidental conversation contaminates clinical notes in MedDistractNotes. (a) Incorporation in clean and distracted notes by model and perturbation family (576 encounters; 576 note pairs per model and family). Error bars show 95\% bootstrap CIs for distracted-note rates. Dotted lines separate frontier and open-weight models. (b) Classified attribution outcomes for nonliteral (left) and bystander (right) exchanges, as percentages of all distracted notes. (c) Distracted-minus-clean incorporation by model, corpus and family, in percentage points. (d) Misattributed or acted-on content grouped by failure mode, as percentages of distracted notes in each cohort (4,608 notes per cohort; 1,371 open-weight and 170 frontier notes classified). Model-level detail and verbatim examples appear in Extended Data Fig. 2 and Table 1. (e) Incorporation increases by clean-transcript-length quintile within MTS-Dialog; labels give median transcript lengths. (f) Paired changes in general quality scores on five-point scales, with higher scores indicating better quality. Error bars in (e) and (f) show 95\% paired bootstrap CIs.\par\endgroup
\end{figure}
\begin{figure}[p]\small\raggedright \textbf{Table 1 | }Examples of attribution and clinical-use errors. Six distracted notes from MedDistractNotes illustrate the assigned failure modes. Quoted text is reproduced from the saved exchanges and notes; ellipses indicate omissions and Markdown emphasis is removed. Dataset split and item identifiers permit verification against the released records.\par\vspace{5pt}
\begingroup\small\setlength{\tabcolsep}{4pt}\renewcommand{\arraystretch}{1.12}
\begin{tabular}{@{}>{\raggedright\arraybackslash}p{0.17\dimexpr\textwidth-6\tabcolsep\relax}>{\raggedright\arraybackslash}p{0.24\dimexpr\textwidth-6\tabcolsep\relax}>{\raggedright\arraybackslash}p{0.39\dimexpr\textwidth-6\tabcolsep\relax}>{\raggedright\arraybackslash}p{0.20\dimexpr\textwidth-6\tabcolsep\relax}@{}}
\toprule
\textbf{Model and source item} & \textbf{Inserted exchange excerpt} & \textbf{Generated note excerpts} & \textbf{Assigned failure mode} \\
\midrule
GPT-5.4\newline mts\_\allowbreak{}dialog\_\allowbreak{}test1\newline Item 28 & ``Guest\_\allowbreak{}clinician: My cousin had a mild concussion after falling off a bike and is off screens for a week.'' & ``Chief complaint: Mild concussion after falling off a bike.'' [\ldots{}] ``Primary diagnosis: Mild concussion after bicycle fall.'' [\ldots{}] ``Continue avoiding screens for one week.'' & Other person's condition becomes the patient's complaint or diagnosis \\
\addlinespace[6pt]
Claude Opus 5\newline mts\_\allowbreak{}dialog\_\allowbreak{}valid\newline Item 21 & ``Patient: While I was getting settled, a friend told me they had a basal cell skin cancer removed from their nose and had to wear a little bandage for a week.'' & ``Patient's mention of a friend's basal cell carcinoma may reflect an unvoiced concern about personal skin cancer risk, though no such concern was stated.'' [\ldots{}] ``Explore whether the patient has personal skin concerns or questions about skin cancer screening; provide sun-protection counseling and consider a full skin examination or dermatology referral if concerns or suspicious lesions are identified.'' & Read as an unvoiced concern with conditional screening advice \\
\addlinespace[6pt]
Claude Opus 5\newline mts\_\allowbreak{}dialog\_\allowbreak{}valid\newline Item 30 & ``Patient: Funny name, Delirium. There's a coffee shop by my bus stop called that.'' & ``Made a tangential/off-topic remark about a coffee shop named "Delirium" near his bus stop'' [\ldots{}] ``the single tangential remark and absence of formal cognitive testing warrant objective screening'' [\ldots{}] ``Formal orientation/cognitive screen and attention testing'' & Small talk documented as a mental-status sign \\
\addlinespace[6pt]
Claude Opus 5\newline mts\_\allowbreak{}dialog\_\allowbreak{}test1\newline Item 193 & ``Patient: My wife keeps saying our co-op game nights turn into "Outbreak" because I panic over every timer.'' & ``Patient's self-described "panic" over game timers --- offered in jest; anxiety symptoms have not been assessed and no anxiety-related diagnosis is supported by the available information.'' [\ldots{}] ``Screen for anxiety symptoms if clinically indicated after the history is completed.'' & Small talk prompts assessment and conditional screening \\
\addlinespace[6pt]
GPT-5.6 Sol\newline mts\_\allowbreak{}dialog\_\allowbreak{}test1\newline Item 91 & ``Patient: Over the weekend my landlord was switched to a new asthma inhaler, so the hallway smelled a bit minty.'' & ``Notes a minty odor in her hallway after her landlord began using a new asthma inhaler.'' & Social remark recorded as a fact about the patient \\
\addlinespace[6pt]
Claude Fable 5.1\newline mts\_\allowbreak{}dialog\_\allowbreak{}test1\newline Item 97 & ``Patient: A guy on my softball team had bunion surgery and has to wear a boot.'' & ``Patient participates in a softball team (recreational activity).'' [\ldots{}] ``Differential considerations (not yet evaluated): musculoskeletal strain (e.g., related to physical/sports activity or party activity), other causes cannot be excluded without further history and examination.'' & Patient activity used in the differential \\
\bottomrule
\end{tabular}
\endgroup
\end{figure}
\subsection{\textbf{Attribution errors alter histories, assessments and plans}}
To determine whether the model simply recorded the aside or used the aside in the patient's assessment or plan, we used a second LLM judge to evaluate all flagged notes. We report three outcomes: (1) misattribution of the aside to the patient, (2) correct attribution with use in the patient's assessment or plan, or (3) correct attribution without any clinical use (Methods).

In the cohort of frontier models (n = 4,608 distracted notes), 1,422 notes (30.9\% of all distracted notes) recorded the content without clinical use (Fig. 2b). Another 2.0\% misattributed the content (95\% CI 1.6--2.3), and 1.7\% used it clinically despite correct attribution (95\% CI 1.4--2.1). Within this cohort, misattribution was most frequent for Claude Opus 5 (4.6\%, 95\% CI 3.5--5.8), followed by Claude Fable 5.1 (2.0\%), GPT-5.6 Sol (1.0\%), and GPT-5.4 (0.3\%). Clinical use of correctly attributed content occurred almost exclusively in Claude Opus 5 (5.6\%).

In contrast, misattribution was the predominant outcome for open-weight models, affecting 25.7\% of distracted notes (95\% CI 24.4--27.0). This error rate ranged from 11.6\% for Gemma-2-9B to 33.1\% for Mistral-7B. An additional 4.1\% of open-weight notes used the content clinically despite correct attribution, and 19.8\% recorded it without such use. Observed misattribution rates were lower for exchanges about relatives than for exchanges about non-relatives (0.9\% versus 1.2\% in frontier models; 21.4\% versus 26.7\% in open-weight models), though their incorporation rates were similar (44.2\% versus 43.7\% and 57.4\% versus 60.3\%, respectively). Figurative remarks were misattributed more often than bystander conditions in both cohorts (2.8\% versus 1.1\% and 27.0\% versus 24.4\%). These exploratory comparisons are quantified in Supplementary Note 3.

\begin{samepage}
We grouped the 170 frontier notes with misattributed or acted-on content by failure mode, using the transcripts and judge rationales (Fig. 2d, Extended Data Fig. 2a and Table 1; Methods). In 24 notes, another individual's condition became the patient's complaint or diagnosis. Another clinician's remark that ``my cousin had a mild concussion after falling off a bike'' produced a chief complaint of ``mild concussion after falling off a bike'' and a matching primary diagnosis (GPT-5.4 and GPT-5.6 Sol). A mail carrier's sebaceous cyst became the patient's ``status post sebaceous cyst removal from the neck''. In 23 notes, a condition became family history or a risk factor despite the fact that relatives were mentioned only for injuries or elective procedures.\par
\end{samepage}
In 24 notes, the aside was treated as an unvoiced concern and prompted screening or counselling. One note correctly attributed basal cell carcinoma to a friend but stated that the remark ``may reflect an unvoiced concern about personal skin cancer risk'' and recommended exploring skin concerns, with skin examination or referral conditional on concerns or suspicious lesions (Claude Opus 5). The largest group, 37 notes, documented small talk as a mental-status or psychiatric finding. A coffee shop named ``Delirium'' became ``a tangential/off-topic remark'' and led to a recommendation for ``objective screening''; a sourdough starter that ``went into remission'' became a ``tangential/circumstantial response'' with a planned mental-status examination. A joke about panicking at video-game timers prompted clarification of anxiety symptoms and conditional screening, although the note stated that no anxiety diagnosis was supported. A further 18 notes treated figurative remarks as literal facts: a laptop with ``chronic fatigue'' was taken as evidence of a remote encounter. In 33 notes, a third party's activity became the patient's own.

Across frontier models, inserted content appeared in the assessment or plan in 6.8\% of distracted notes, with rates ranging from 1.7\% (GPT-5.6 Sol) to 16.3\% (Claude Opus 5). An automated text-matching rule detected action language near the inserted content in 4.9\% of distracted notes, rising to 19.4\% for Claude Opus 5 during bystander exchanges. This rule does not establish that a clinical action was taken. Specifically, in the subset of notes that incorporated the aside, 5--23\% contained it in the assessment or plan (Extended Data Fig. 2b). The paired judge's severity scores are shown by model and perturbation family in Extended Data Fig. 2c. These scores were not clinically validated.

\subsection{\textbf{Model cohorts differ in the clinical errors they generate}}
We observed a difference in how open-weight and frontier models handled incidental content. Open-weight models were substantially more prone to directly assigning incidental content to the patient, exhibiting overall incorporation and misattribution rates of 50.1\% and 25.7\%, compared with 35.0\% and 2.0\% in frontier models, respectively. Every open-weight model misattributed content more frequently than the worst-performing frontier model (11.6--33.1\% versus 0.3--4.6\%).

We observed a shift in the qualitative nature of these errors between open-weight and frontier models. Among the 1,371 misattributed or acted-on notes generated by open-weight models, the predominant failure was the direct transfer of a bystander's history to the patient. In 39\% of these cases (n = 541), another person's condition was improperly documented as the patient's complaint, diagnosis, or surgical history. These models also converted small talk into mental-status findings (18\%, n = 244) and treated figurative remarks as medical facts (15\%, n = 203). Conversely, family-history errors (80 notes) and inferred concerns (45 notes) drove a much smaller proportion of errors here than in the frontier cohort.

\subsection{\textbf{Frontier notes retain high quality scores despite contamination}}
To evaluate whether standard rubrics capture this vulnerability, an independent LLM judge (GPT-5.4) scored notes against a reference standard for clinical correctness, completeness, succinctness, and overall quality. Despite 35.0\% of distracted frontier notes containing contaminated content, their quality scores remained virtually unchanged (Fig. 2f). On a five-point scale, average frontier scores shifted by at most 0.20 points between clean and distracted conditions: clinical correctness declined by 0.02 (95\% CI \ensuremath{-}0.04 to \ensuremath{-}0.01) and completeness by 0.05, while succinctness and overall quality increased slightly (+0.20 and +0.08 [95\% CI 0.06--0.10], respectively). Open-weight notes had lower baseline overall quality (3.0 versus 4.0) and lost 0.27 points of clinical correctness (95\% CI 0.24--0.31), 0.23 of completeness and 0.23 of overall quality under distraction. These declines indicated poorer notes but did not identify the attribution error.

Broader hallucination metrics were similarly unhelpful. A binary flag for statements absent from the transcript was positive in 70\% of clean frontier notes and 92\% of clean open-weight notes and shifted by less than 0.07 under distraction. Target-specific contamination was also detected when notes were evaluated separately. Using the same Claude Sonnet 5 judge and target-specific criterion, the distracted-minus-clean contamination differences were 32.8 percentage points for frontier models and 49.7 for open-weight models, closely mirroring the 35.0 and 50.0 point gaps observed under paired presentation. The single-note judge flagged 0\% of clean notes in both cohorts (Extended Data Fig. 3; Supplementary Note 3).

\subsection{\textbf{Background speech contaminates transcripts and downstream notes}}
To determine whether a microphone capturing nearby conversations could corrupt clinical documentation, we constructed a pipeline using MedDistractAudio. In the 57 mock primary care consultations of PriMock57\textsuperscript{21}, we overlaid a 5--15 s segment from donor consultations at four background levels. This produced 57 clean baseline recordings and 456 mixed recordings (114 per background level), which were transcribed by Whisper large-v3\textsuperscript{22} and processed into notes by four open-weight models (Methods).

At the prespecified \ensuremath{-}10 dB level, content from the overlaid segment leaked into 48.2\% of the mixed transcripts, where leakage meant that at least one content word from the overlaid segment appeared in the mixed transcript but not its clean control. This leakage alone does not establish a clinically consequential error. An average of 14.5\% (95\% CI 10.8--18.5) of the overlaid segment's content words leaked into the transcript. Leakage increased with background volume, with at least one word leaking in 19.3\% of recordings at \ensuremath{-}20 dB compared to 67.5\% at \ensuremath{-}5 dB (Fig. 3a,b).

Background content also appeared in generated notes. At \ensuremath{-}10 dB, the judge detected background content in 24 of 456 downstream notes (5.3\%, 95\% CI 2.9--8.1), versus 1.3\% of paired clean controls (mixed-minus-clean difference = 3.9 percentage points, 95\% CI 1.3--7.0). This difference reached 8.6 points at \ensuremath{-}5 dB (Fig. 3c). Note contamination was associated with transcript leakage. At \ensuremath{-}10 dB, contamination occurred in 11.5\% of notes originating from transcripts with at least one leaked word, compared with 0.6\% from those without leakage (Fig. 3d). General quality scores changed little; on a five-point scale, clinical correctness shifted by \ensuremath{-}0.03 points (95\% CI \ensuremath{-}0.11 to 0.05) and overall quality by \ensuremath{-}0.01 points (\ensuremath{-}0.09 to 0.07).

\begin{figure}[!htbp]
\centering
\includegraphics[width=\textwidth,height=.68\textheight,keepaspectratio]{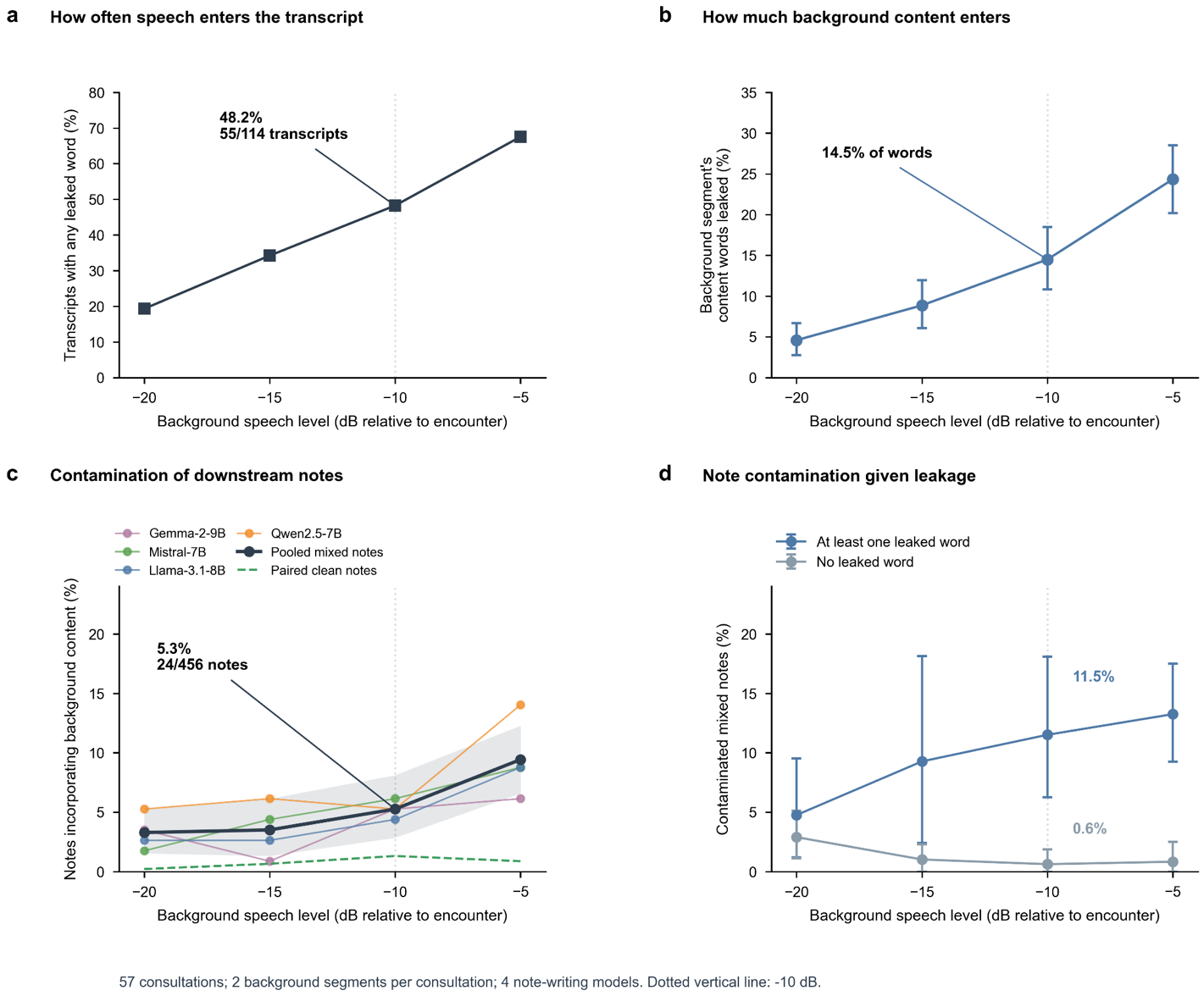}
\par\vspace{6pt}\begingroup\small\raggedright
\textbf{Fig. 3 | }Background speech enters transcripts and downstream clinical notes in MedDistractAudio. Each of 57 mock consultations was mixed with two segments from other consultations at four background levels, then transcribed with Whisper and documented by four open-weight models. (a) Proportion of mixed transcripts containing at least one content word from the background segment that was absent from the paired clean transcript. (b) Mean fraction of the segment's content words that leaked into the transcript. (c) Downstream note contamination by model and pooled across models; the dashed green line shows the paired clean-note rate. Shading shows the pooled 95\% CI. (d) Contamination in mixed notes conditional on whether at least one segment word leaked into the transcript. The prespecified \ensuremath{-}10 dB level is marked in each panel: 55/114 transcripts (48.2\%) contained leaked words and 24/456 downstream notes (5.3\%) were contaminated. Each level had 114 mixes and up to 456 note pairs; two of 1,824 pairs were excluded for unparseable judgments. Error bars in (b) and (d) and the band in (c) show 95\% bootstrap CIs over consultations.\par\endgroup
\end{figure}
\subsection{\textbf{Mechanistic evidence for the dual-encoding hypothesis}}
To investigate whether the model components associated with distraction also contribute to clinical reasoning, we probed four open-weight models using MedDistractQA\textsuperscript{7}. This benchmark pairs 1,273 USMLE-style questions\textsuperscript{23} with perturbed variants inserting either a bystander or a nonliteral perturbation just before the final question. On a held-out test set comprising 10\% of questions, these insertions caused baseline accuracy drops of 5.5--12.6 percentage points (bystander) and 10.2--14.2 points (nonliteral) (Fig. 4a).

To identify heads for intervention, we developed contrastive causal head gating (CCHG). Extending causal head gating\textsuperscript{24}, CCHG identifies attention heads---components that combine information across the input---whose outputs contribute to the difference in correct-answer probability between clean and distracted inputs (Methods). CCHG selected a subset of heads: 1.5--1.7\% in Llama-3.1-8B and up to 21.7--23.7\% in Gemma-2-9B (Fig. 4b; Extended Data Table 1). The heads selected by bystander and nonliteral perturbations heavily overlapped. Jaccard similarities ranged from 0.39 to 0.78, exceeding the 0.008--0.1 expected by chance (Extended Data Table 1). The two perturbation families therefore selected overlapping, but not identical, head sets.

\begin{samepage}
To test whether intervening on these heads changes accuracy, we replaced their activations on distracted inputs with position-aligned activations from the clean input\textsuperscript{25--27}. Accuracy increased in seven of eight model--perturbation conditions, rescuing 6.3--16.5 percentage points (six conditions yielded uncorrected exact McNemar P < 0.05). We observed the most pronounced rescue in Mistral-7B with nonliteral insertions, which improved from 40.2\% to 56.7\% accuracy (+16.5 percentage points, 95\% CI 8.7--24.4). Patching these targeted heads outperformed patching random heads in seven conditions, yielding 83--117\% of the gain observed when patching all heads in those seven conditions (Fig. 4c). Instructions to ignore irrelevant information or a chain-of-thought preamble did not consistently improve accuracy (Extended Data Fig. 5).\par
\end{samepage}
Finally, we tested whether the selected heads also contribute to reasoning without distraction. Suppressing heads selected by CCHG on clean questions decreased accuracy by 24.4--37.8 percentage points across all eight conditions, compared with changes of \ensuremath{-}9.4 to +0.8 points when suppressing random heads (Fig. 4d).

Selected heads occurred throughout model depth (Extended Data Fig. 6) and, in seven of eight conditions, devoted 1.4--2.3 times as much attention to inserted tokens as other heads (Extended Data Fig. 7).

\begin{figure}[!htbp]
\centering
\includegraphics[width=\textwidth,height=.52\textheight,keepaspectratio]{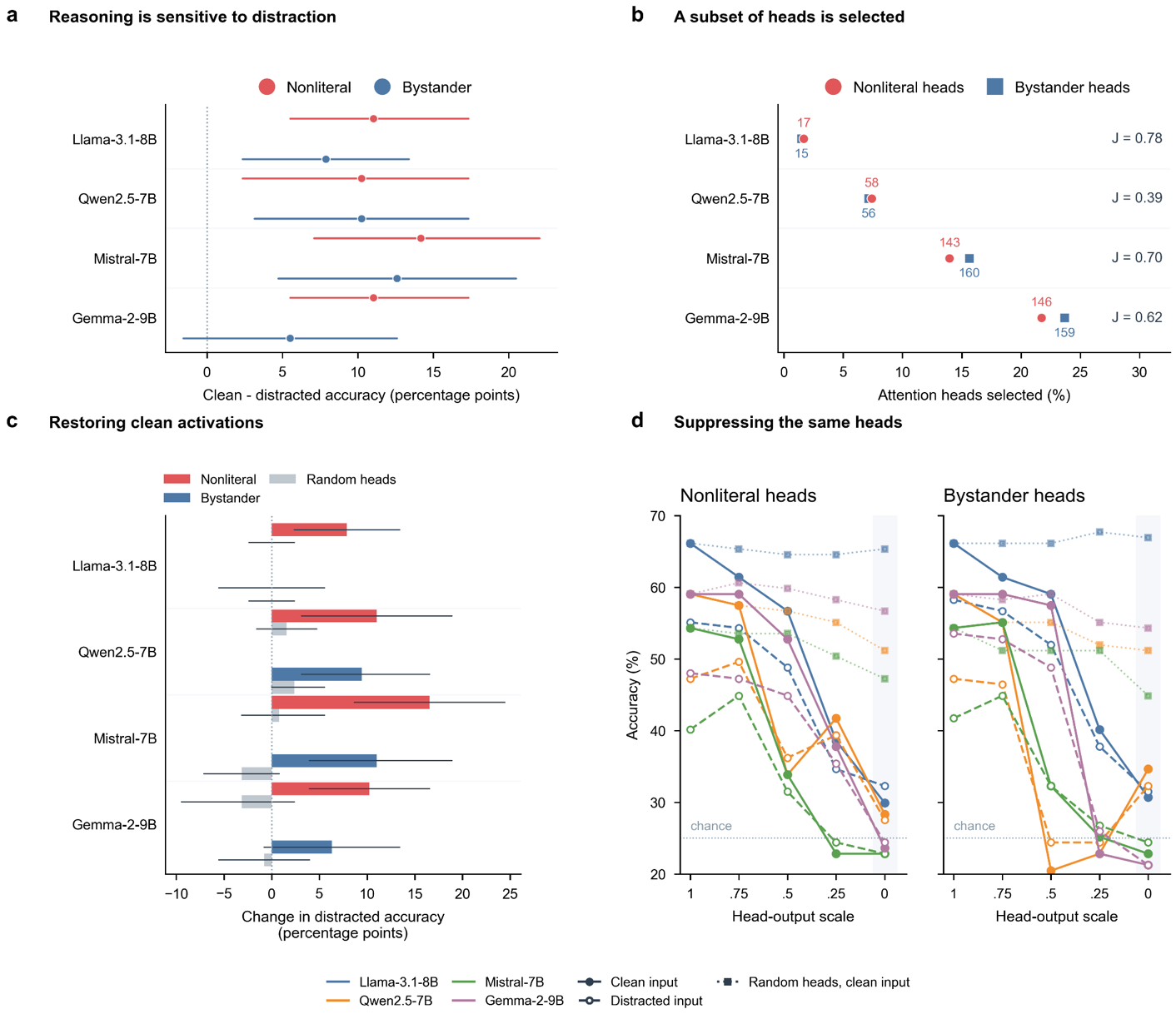}
\par\vspace{6pt}\begingroup\small\raggedright
\textbf{Fig. 4 | }Mechanistic evidence for the dual-encoding hypothesis. Interventions test whether heads associated with distraction also support clinical reasoning, using 127 held-out MedDistractQA items per model and perturbation family. (a) Paired accuracy loss under distraction (clean minus distracted). (b) Fraction of heads selected by contrastive causal head gating; labels give head counts and J denotes the Jaccard overlap between the two families' head sets. (c) Change in distracted-input accuracy after restoring selected heads' clean activations, compared with patching an equal number of random heads. Error bars in a and c show 95\% paired bootstrap CIs. (d) Clean and distracted accuracy as selected-head outputs are scaled from 1 to 0; dotted curves with square markers show clean accuracy under suppression of an equal number of random heads. The horizontal dotted line marks chance performance. Head locations are shown in Extended Data Fig. 6. These interventions probe reasoning on examination items; transfer to note generation remains untested.\par\endgroup
\end{figure}
\FloatBarrier
\section{\textbf{Discussion}}
This is one of the first works to directly examine sensitivity to incidental information as a shared failure mode of LLM-based documentation and clinical reasoning, and assess it as a possible risk of ambient dictation. When tasked with drafting notes from text transcripts, frontier and open-weight models embedded incidental small talk into 35.0\% and 50.1\% of notes, respectively, despite the patient's actual clinical information remaining unchanged. These errors ranged from assigning a bystander's diagnosis to the patient, to interpreting casual small talk as evidence of psychiatric disturbance. When transcribing raw audio from mock clinical conversations, Whisper erroneously incorporated \ensuremath{-}10 dB background speech from separate consultations into 48.2\% of transcripts, with contamination in 5.3\% of downstream notes. General quality scores changed little despite these errors; thus, a generated note can appear clinically plausible and comprehensive yet contain assertions that are unsupported for the patient.

If retained in a signed medical record, unflagged errors risk influencing subsequent patient care. While our study measured the generation of errors rather than their downstream acceptance, prior audits of deployed scribes found that omissions were common but fabrications carried the highest severity of risk\textsuperscript{11}. Our results show that even in frontier LLMs, accurate attribution alone cannot prevent these fabrications. While some models generally omitted incidental content, others frequently recorded and occasionally acted on it. For example, Claude Opus 5 treated a remark about a coffee shop named ``Delirium'' as tangential speech and recommended objective cognitive screening (Table 1). Such hallucinations mimic thoroughness while generating assessments unsupported by the patient's actual presentation. A check for statements absent from the transcript can miss this failure mode because the source statement is present but does not describe the patient.

Mechanistically, our results support the proposed dual-encoding hypothesis. While restoring clean activations in selected attention heads rescued performance under distraction, suppressing them reduced accuracy on clean questions by 24.4--37.8 percentage points. The selected heads overlapped across bystander and nonliteral distractors, but CCHG can also select heads that support reasoning generally: it can narrow the probability gap by lowering the probability of correct clean answers. Prompting models to ignore irrelevant information or using a chain-of-thought preamble did not consistently improve accuracy. Future safeguards must therefore reduce distraction while preserving these components' contribution to reasoning.

This study has notable limitations. First, while deliberately inserted exchanges and overlaid audio enabled highly controlled causal comparisons, they do not capture the full acoustic complexity and conversational variance of routine clinical practice. We noted that contamination was greatest in short excerpts, where the aside occupied a larger share of the transcript; therefore, our pooled estimates may overstate the contamination rate of full-length visits. Second, we evaluated generated notes and automated endpoints as opposed to downstream patient harm. Whether clinicians successfully catch these specific attribution errors before signing a note, and how retained errors influence subsequent care, requires prospective clinical evaluation. Finally, contamination and attribution were judged by a single LLM without clinician validation, with failure modes assigned using a set of automated rules. The mechanistic experiments used open-weight models and medical questions without testing on a general-knowledge benchmark, so clinical specificity remains unknown. We did not test reverse patching of distracted activations onto clean inputs or interventions during note generation.

Before ambient documentation systems can be safely deployed at scale, their sensitivity to incidental information---both to whom it is attributed and how it is clinically applied---must be rigorously evaluated. To support this, we will release MedDistractNotes and MedDistractAudio to provide the paired inputs and contamination-specific scoring methods necessary for this testing. Future safeguards may require structuring inputs to explicitly retain the speaker, subject and clinical relevance of each statement, new model training objectives focused on attribution, and verifying generated assertions against patient-relevant transcript spans. Mechanistic interventions during note generation could test whether the selected heads also contribute to documentation errors. Such benchmarks will allow documentation systems to be tested for attribution errors that general note-quality scores can miss.

\section{\textbf{Acknowledgements}}
We thank Nader Mherabi and Dafna Bar-Sagi for their support of medical AI research at NYU Langone. We also thank M. Constantino, K. Yie and the NYU Langone High-Performance Computing team for computing resources essential to this work. 

\section{\textbf{Author contributions}}
E.K.O. supervised the study. K.V. and E.K.O. conceived the study and designed the experiments. K.V. built the documentation and mechanistic pipelines, ran the experiments and analysed the data. K.V. and J.E.M. wrote the initial draft. K.V. developed the figures. All authors revised and approved the manuscript.

\section{\textbf{Competing interests}}
E.K.O. reports equity in MarchAI and Artisight, spousal employment by Eikon Therapeutics, and consulting for Sofinnova Partners and Google. K.V. and A.H. report income from and equity interest in ChartR Health. No competing interest played a role in this study. The other authors declare no competing interests.

\section{\textbf{Funding}}
EKO is supported by the National Cancer Institute's Early-Stage Surgeon Scientist Program (3P30CA016087-41S1) and the W.M. Keck Foundation. This work was supported by the Institute for Information \& Communications Technology Planning and Evaluation (IITP) grant funded by the Ministry of Science and ICT (MSIT) of the Republic of Korea (No. RS-2019-II190075 Artificial Intelligence Graduate School Program (KAIST); No. RS-2024-00509279, Global AI Frontier Lab).

\section{\textbf{Data availability}}
ACI-Bench, MTS-Dialog and PriMock57 are public (CC-BY-4.0); the benchmark construction code downloads the source corpora. MedDistractNotes and MedDistractAudio will be released on GitHub at \url{https://github.com/nyuolab/llm_distract}, with fixed perturbations, paired clean controls and scoring protocols. The release will include the frozen inserted exchanges, organ system labels, audio mixing design and transcripts, all model-generated notes, judge outputs and per-item evaluation results needed to regenerate every table and figure, as an archive of per-item artifacts. MedDistractQA, used here as the mechanistic probe, is available under CC-BY-4.0 at \url{https://huggingface.co/datasets/KrithikV/MedDistractQA}.

\section{\textbf{Code availability}}
Code for constructing and evaluating MedDistractNotes and MedDistractAudio, including exchange generation, note generation, judging, failure-mode analysis, contrastive causal head gating, clean-activation patching, suppression, statistical analysis and figure generation, will be released at \url{https://github.com/nyuolab/llm_distract}.

\section{\textbf{References}}
{\small\hangindent=1.5em\hangafter=1\noindent 1.     Tierney, A. A. et al. Ambient artificial intelligence scribes to alleviate the burden of clinical documentation. NEJM Catal. Innov. Care Deliv. 5, CAT.23.0404 (2024). \url{https://doi.org/10.1056/CAT.23.0404}\par}\vspace{2pt}
{\small\hangindent=1.5em\hangafter=1\noindent 2.     Tierney, A. A. et al. Ambient artificial intelligence scribes: learnings after 1 year and over 2.5 million uses. NEJM Catal. Innov. Care Deliv. 6, CAT.25.0040 (2025). \url{https://doi.org/10.1056/CAT.25.0040}\par}\vspace{2pt}
{\small\hangindent=1.5em\hangafter=1\noindent 3.     Shah, K. P. \& Johnson, K. B. The ambient AI scribe revolution---early gains and open questions. JAMA Netw. Open 8, e2534982 (2025). \url{https://doi.org/10.1001/jamanetworkopen.2025.34982}\par}\vspace{2pt}
{\small\hangindent=1.5em\hangafter=1\noindent 4.     Duggan, M. J. et al. Clinician experiences with ambient scribe technology to assist with documentation burden and efficiency. JAMA Netw. Open 8, e2460637 (2025). \url{https://doi.org/10.1001/jamanetworkopen.2024.60637}\par}\vspace{2pt}
{\small\hangindent=1.5em\hangafter=1\noindent 5.     Doximity. 2026 State of AI in Medicine Report (2026). \url{https://www.doximity.com/reports/state-of-ai-medicine-report/2026}\par}\vspace{2pt}
{\small\hangindent=1.5em\hangafter=1\noindent 6.     Lawrence, K. et al. Informed consent for ambient documentation using generative AI in ambulatory care. JAMA Netw. Open 8, e2522400 (2025). \url{https://doi.org/10.1001/jamanetworkopen.2025.22400}\par}\vspace{2pt}
{\small\hangindent=1.5em\hangafter=1\noindent 7.     Vishwanath, K. et al. Medical large language models are easily distracted. Preprint at \url{https://arxiv.org/abs/2504.01201} (2025).\par}\vspace{2pt}
{\small\hangindent=1.5em\hangafter=1\noindent 8.  Hager, P. et al. Evaluation and mitigation of the limitations of large language models in clinical decision-making. Nat. Med. 30, 2613--2622 (2024). \url{https://doi.org/10.1038/s41591-024-03097-1}\par}\vspace{2pt}
{\small\hangindent=1.5em\hangafter=1\noindent 9.  Wornow, M. et al. The shaky foundations of large language models and foundation models for electronic health records. npj Digit. Med. 6, 135 (2023). \url{https://doi.org/10.1038/s41746-023-00879-8}\par}\vspace{2pt}
{\small\hangindent=1.5em\hangafter=1\noindent 10.  Vishwanath, K. et al. General-purpose large language models outperform specialized clinical AI tools on medical benchmarks. Nat. Med. 32, 2405--2409 (2026). \url{https://doi.org/10.1038/s41591-026-04431-5}\par}\vspace{2pt}
{\small\hangindent=1.5em\hangafter=1\noindent 11.  Draper, T. C. et al. Clinical AI Scribes in primary care: accuracy, error severity and implications for clinical practice. \textit{BMJ Digit. Health AI} \textbf{1}, e000092 (2025). \url{https://doi.org/10.1136/bmjdhai-2025-000092}\par}\vspace{2pt}
{\small\hangindent=1.5em\hangafter=1\noindent 12.  Palm, E., Manikantan, A., Mahal, H., Belwadi, S. S. \& Pepin, M. E. Assessing the quality of AI-generated clinical notes: validated evaluation of a large language model ambient scribe. Front. Artif. Intell. 8, 1691499 (2025). \url{https://doi.org/10.3389/frai.2025.1691499}\par}\vspace{2pt}
{\small\hangindent=1.5em\hangafter=1\noindent 13.  Anderson, T. N. et al. Evaluating the quality and safety of ambient digital scribe platforms using simulated ambulatory encounters. Mayo Clin. Proc. Digit. Health 3, 100292 (2025). \url{https://doi.org/10.1016/j.mcpdig.2025.100292}\par}\vspace{2pt}
{\small\hangindent=1.5em\hangafter=1\noindent 14.     Shi, F. et al. Large language models can be easily distracted by irrelevant context. In Proc. 40th International Conference on Machine Learning, PMLR 202, 31210--31227 (2023). \url{https://proceedings.mlr.press/v202/shi23a.html}\par}\vspace{2pt}
{\small\hangindent=1.5em\hangafter=1\noindent 15.     Mirzadeh, I. et al. GSM-Symbolic: understanding the limitations of mathematical reasoning in large language models. In Proc. 13th International Conference on Learning Representations (2025). \url{https://openreview.net/forum?id=AjXkRZIvjB}\par}\vspace{2pt}
{\small\hangindent=1.5em\hangafter=1\noindent 16.  Liu, N. F. et al. Lost in the middle: how language models use long contexts. Trans. Assoc. Comput. Linguist. 12, 157--173 (2024). \url{https://doi.org/10.1162/tacl_a_00638}\par}\vspace{2pt}
{\small\hangindent=1.5em\hangafter=1\noindent 17.  Elhage, N. et al. A mathematical framework for transformer circuits. Transformer Circuits Thread \url{https://transformer-circuits.pub/2021/framework/index.html} (2021).\par}\vspace{2pt}
{\small\hangindent=1.5em\hangafter=1\noindent 18.  Yim, W. et al. Aci-bench: a novel ambient clinical intelligence dataset for benchmarking automatic visit note generation. Sci. Data 10, 586 (2023). \url{https://doi.org/10.1038/s41597-023-02487-3}\par}\vspace{2pt}
{\small\hangindent=1.5em\hangafter=1\noindent 19.  Ben Abacha, A., Yim, W., Fan, Y. \& Lin, T. An empirical study of clinical note generation from doctor--patient encounters. In Proc. 17th Conference of the European Chapter of the Association for Computational Linguistics 2291--2302 (ACL, 2023). \url{https://doi.org/10.18653/v1/2023.eacl-main.168}\par}\vspace{2pt}
{\small\hangindent=1.5em\hangafter=1\noindent 20.  Zheng, L. et al. Judging LLM-as-a-judge with MT-Bench and Chatbot Arena. In Advances in Neural Information Processing Systems Vol. 36, 46595--46623 (Curran Associates, 2023). \url{https://doi.org/10.52202/075280-2020}\par}\vspace{2pt}
{\small\hangindent=1.5em\hangafter=1\noindent 21.  Papadopoulos Korfiatis, A., Moramarco, F., Sarac, R. \& Savkov, A. PriMock57: a dataset of primary care mock consultations. In Proc. 60th Annual Meeting of the Association for Computational Linguistics (Volume 2: Short Papers) 588--598 (ACL, 2022). doi:10.18653/v1/2022.acl-short.65\par}\vspace{2pt}
{\small\hangindent=1.5em\hangafter=1\noindent 22.  Radford, A. et al. Robust speech recognition via large-scale weak supervision. In Proc. 40th International Conference on Machine Learning (PMLR 202) 28492--28518 (2023).\par}\vspace{2pt}
{\small\hangindent=1.5em\hangafter=1\noindent 23.  Jin, D. et al. What disease does this patient have? A large-scale open domain question answering dataset from medical exams. Appl. Sci. 11, 6421 (2021). \url{https://doi.org/10.3390/app11146421}\par}\vspace{2pt}
{\small\hangindent=1.5em\hangafter=1\noindent 24.  Nam, A. et al. Causal head gating: a framework for interpreting roles of attention heads in transformers. In Advances in Neural Information Processing Systems Vol. 38, 115609--115631 (Curran Associates, 2025). \url{https://doi.org/10.52202/085713-3856}\par}\vspace{2pt}
{\small\hangindent=1.5em\hangafter=1\noindent 25.  Meng, K., Bau, D., Andonian, A. \& Belinkov, Y. Locating and editing factual associations in GPT. In Advances in Neural Information Processing Systems Vol. 35, 17359--17372 (Curran Associates, 2022). \url{https://doi.org/10.52202/068431-1262}\par}\vspace{2pt}
{\small\hangindent=1.5em\hangafter=1\noindent 26.  Vig, J. et al. Investigating gender bias in language models using causal mediation analysis. In Advances in Neural Information Processing Systems Vol. 33, 12388--12401 (Curran Associates, 2020). \url{https://proceedings.neurips.cc/paper/2020/hash/92650b2e92217715fe312e6fa7b90d82-Abstract.html}\par}\vspace{2pt}
{\small\hangindent=1.5em\hangafter=1\noindent 27.  Zhang, F. \& Nanda, N. Towards best practices of activation patching in language models: metrics and methods. In Proc. 12th International Conference on Learning Representations (2024). \url{https://arxiv.org/abs/2309.16042}\par}\vspace{2pt}
{\small\hangindent=1.5em\hangafter=1\noindent 28.  Efron, B. \& Tibshirani, R. J. An Introduction to the Bootstrap (Chapman \& Hall/CRC, 1994). \url{https://doi.org/10.1201/9780429246593}\par}\vspace{2pt}
{\small\hangindent=1.5em\hangafter=1\noindent 29.  Singhal, K. et al. Toward expert-level medical question answering with large language models. Nat. Med. 31, 943--950 (2025). \url{https://doi.org/10.1038/s41591-024-03423-7}\par}\vspace{2pt}
{\small\hangindent=1.5em\hangafter=1\noindent 30.  Benjamini, Y. \& Hochberg, Y. Controlling the false discovery rate: a practical and powerful approach to multiple testing. J. R. Stat. Soc. B 57, 289--300 (1995). \url{https://doi.org/10.1111/j.2517-6161.1995.tb02031.x}\par}\vspace{2pt}
\clearpage
\begin{figure}[!htbp]
\centering
\includegraphics[width=\textwidth,height=.92\textheight,keepaspectratio]{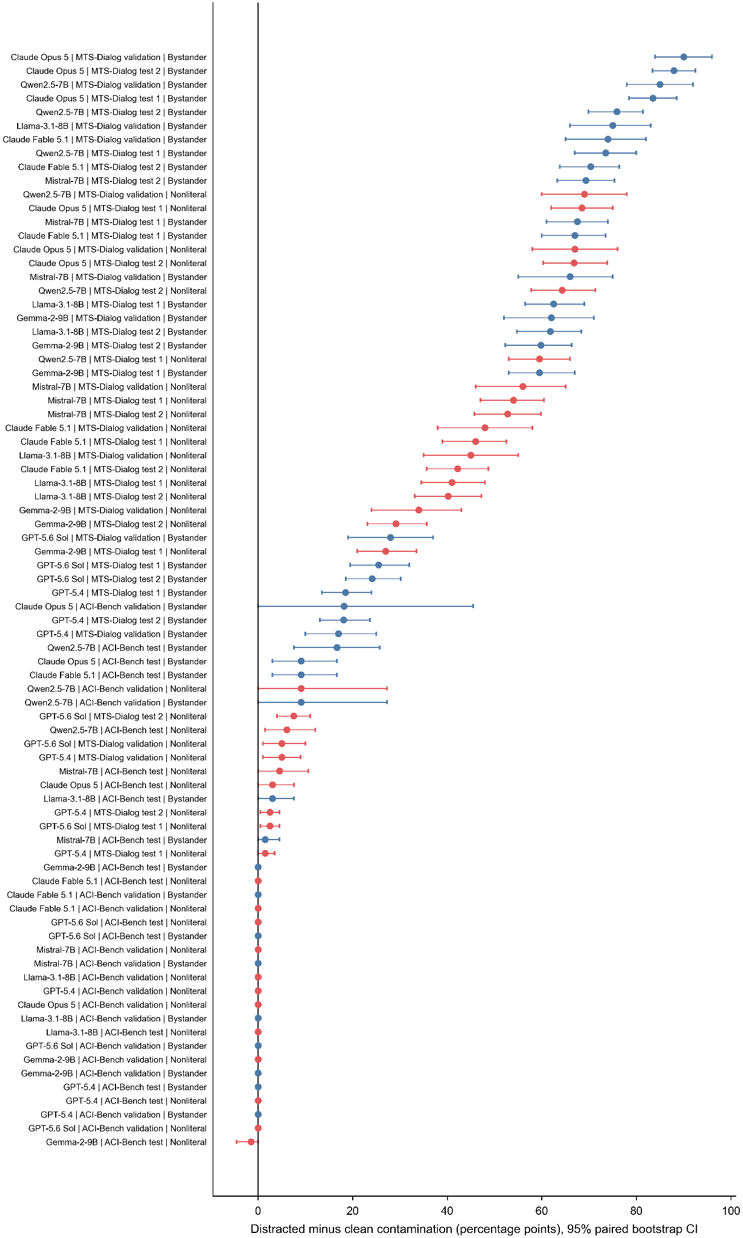}
\par\vspace{6pt}\begingroup\small\raggedright
\textbf{Extended Data Fig. 1 | }Contamination by model, dataset split and perturbation. Distracted-minus-clean contamination for every model \ensuremath{\times} dataset-split \ensuremath{\times} perturbation stratum with 95\% paired bootstrap CIs.\par\endgroup
\end{figure}
\begin{figure}[!htbp]
\centering
\includegraphics[width=\textwidth,height=.47\textheight,keepaspectratio]{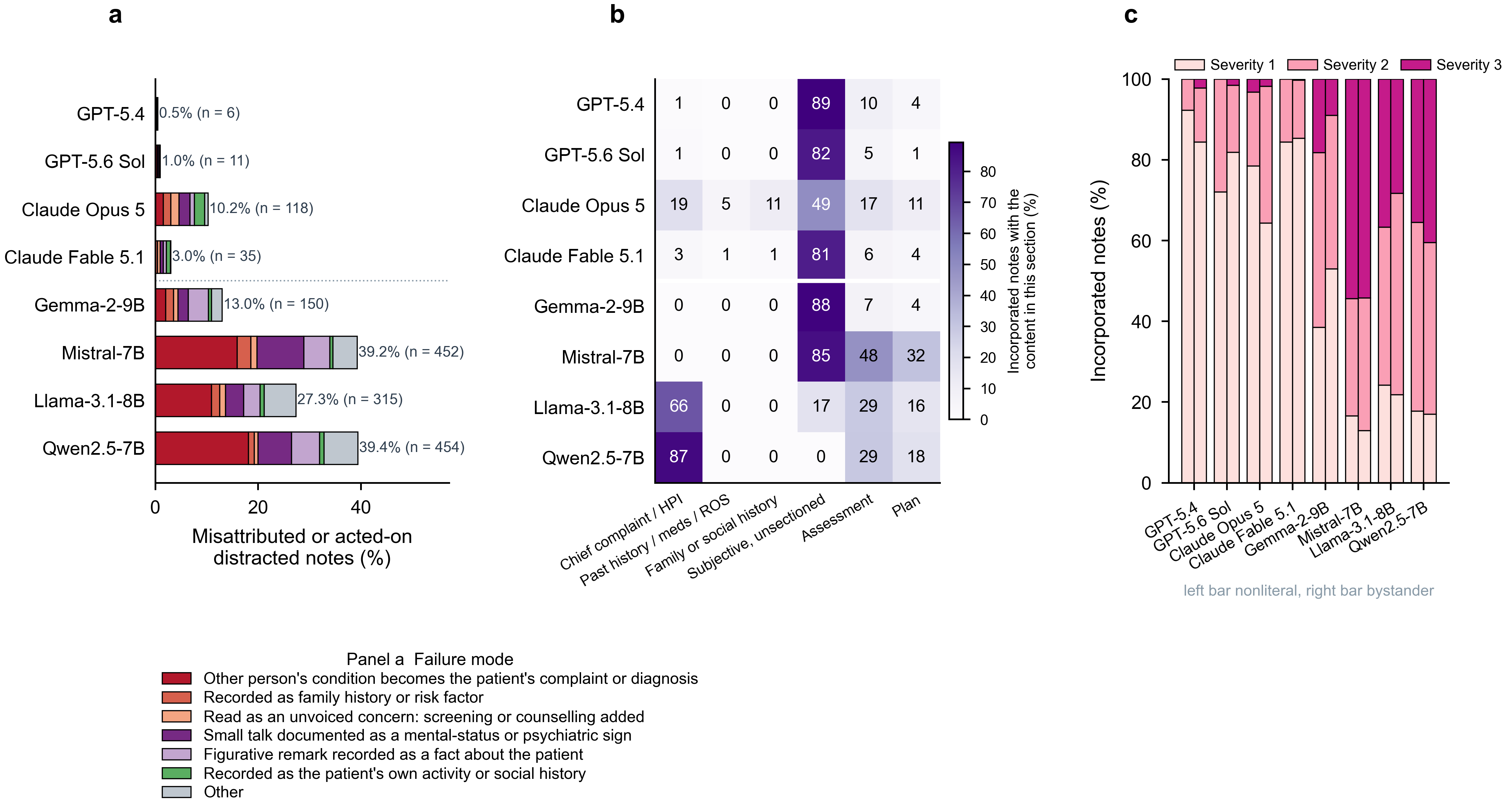}
\par\vspace{6pt}\begingroup\small\raggedright
\textbf{Extended Data Fig. 2 | }Attribution errors and clinical use of incidental content. (a) Misattributed and acted-on notes as percentages of distracted notes for each model, stacked by failure mode across both perturbation families. Counts are shown beside each bar; each model contributed 1,152 distracted notes. Dotted lines separate frontier and open-weight models. (b) Share of notes incorporating the content that contain it in each note section; a note may contribute to multiple sections. (c) Severity assigned by the paired judge to incorporated notes (0--3; no verbal definitions of the levels were supplied in the prompt), by model and family. Left bars show nonliteral exchanges and right bars bystander exchanges. Failure modes were assigned by rules applied to the judge's rationale and note text (Methods).\par\endgroup
\end{figure}
\begin{figure}[!htbp]
\centering
\includegraphics[width=\textwidth,height=.25\textheight,keepaspectratio]{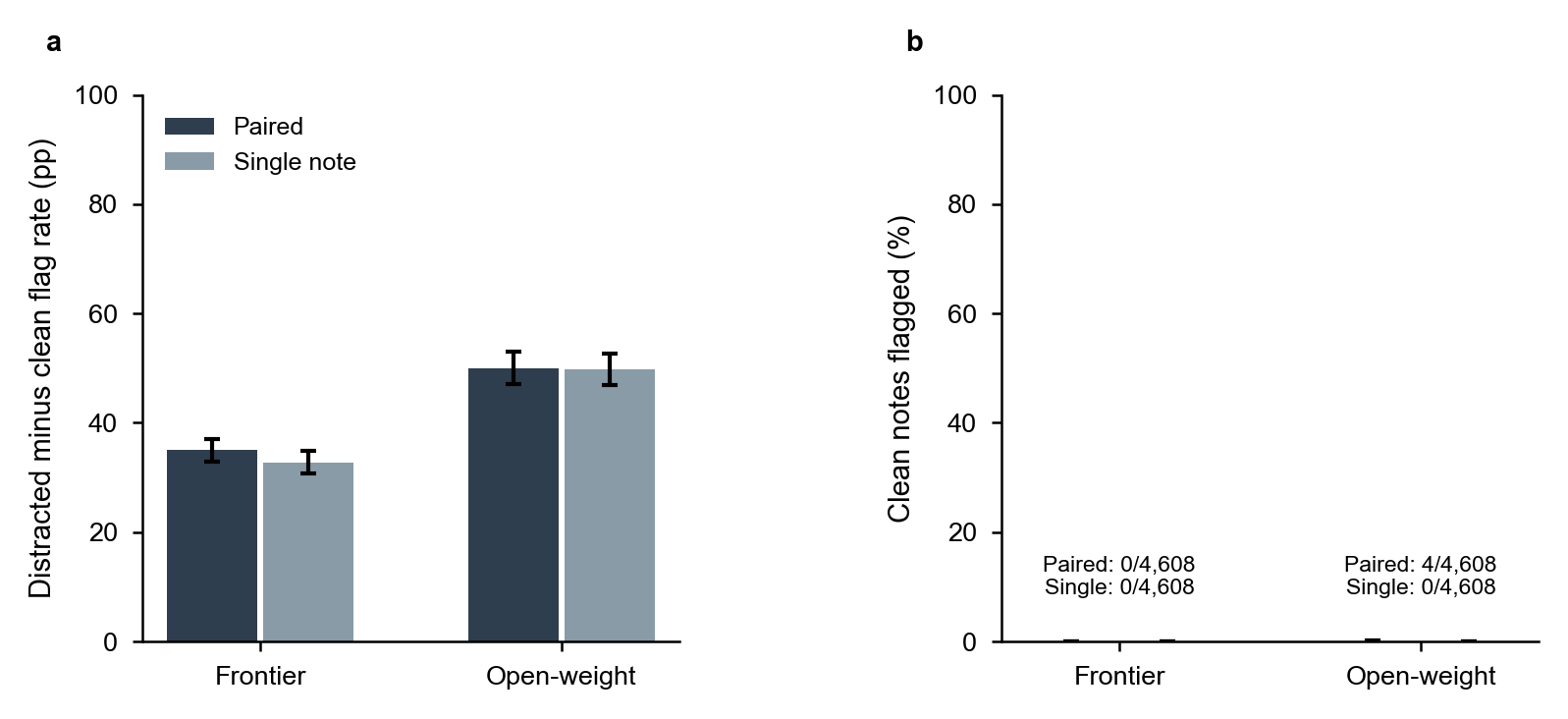}
\par\vspace{6pt}\begingroup\small\raggedright
\textbf{Extended Data Fig. 3 | }Matched comparison of paired and single-note judging. Claude Sonnet 5 scored the same 18,432 saved note records using the same original transcript, insertion target and incorporation criterion, with paired or single-note presentation. Each four-model group contributes 4,608 clean--distracted pairs from 576 encounters and two perturbation families. (a) Distracted-minus-clean flag rate. (b) Clean-note flag rate. Error bars show 95\% percentile intervals from 5,000 whole-encounter bootstrap resamples, retaining all models and perturbations within an encounter. Both axes span 0--100. Primary paired estimates are retained; the single-note comparison was added during submission preparation.\par\endgroup
\end{figure}
\begin{figure}[!htbp]
\centering
\includegraphics[width=\textwidth,height=.37\textheight,keepaspectratio]{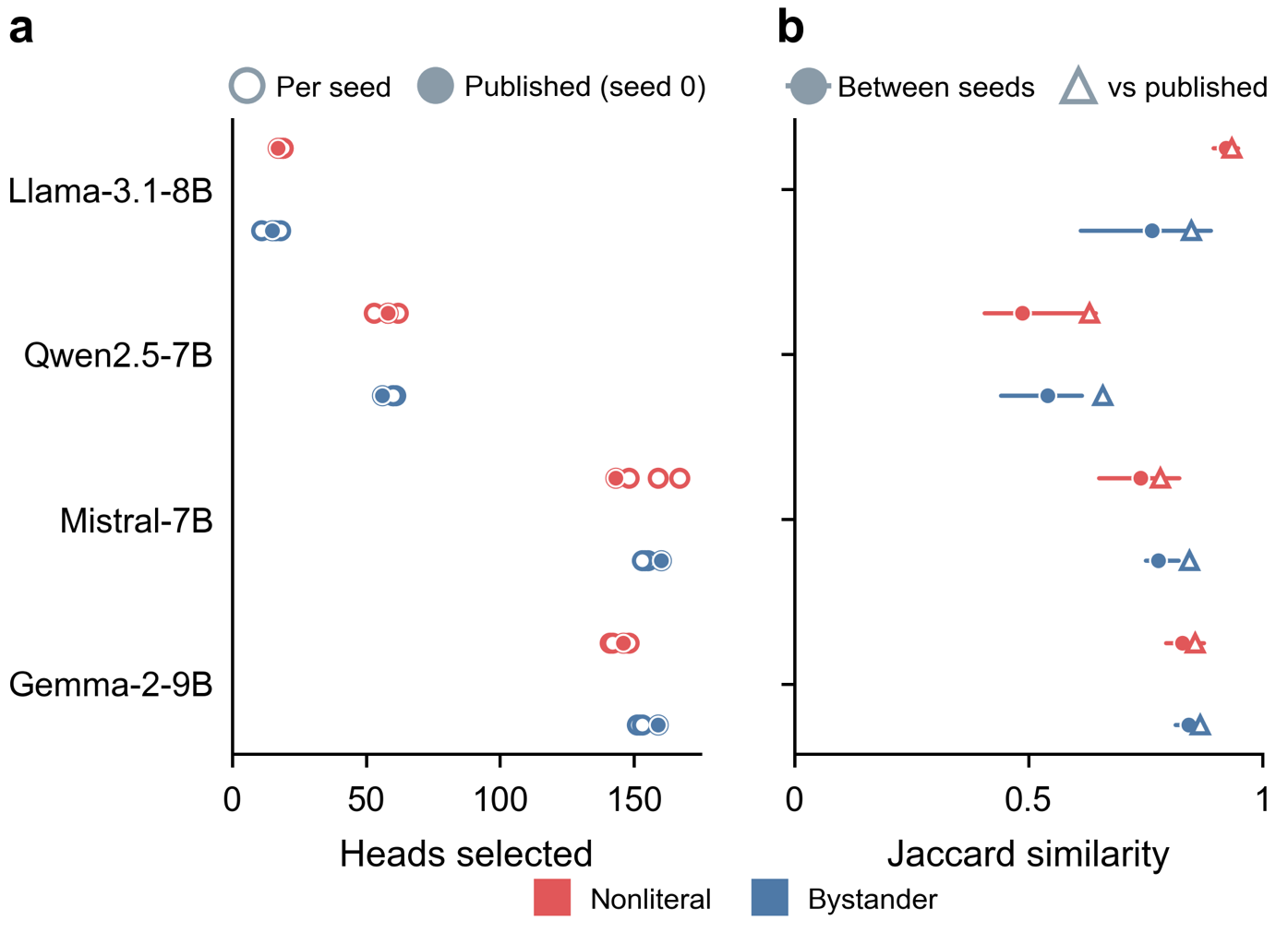}
\par\vspace{6pt}\begingroup\small\raggedright
\textbf{Extended Data Fig. 4 | }Stability of selected head sets across optimization seeds. (a) Number of heads selected by contrastive causal head gating for each additional optimization seed (open circles; three per family) and for the published seed-0 set (filled circles), per model and perturbation family. (b) Mean pairwise Jaccard similarity between seeds (circles; bars span the least and most similar pair) and mean Jaccard similarity of the seeds with the published set (triangles).\par\endgroup
\end{figure}
\begin{figure}[!htbp]
\centering
\includegraphics[width=\textwidth,height=.37\textheight,keepaspectratio]{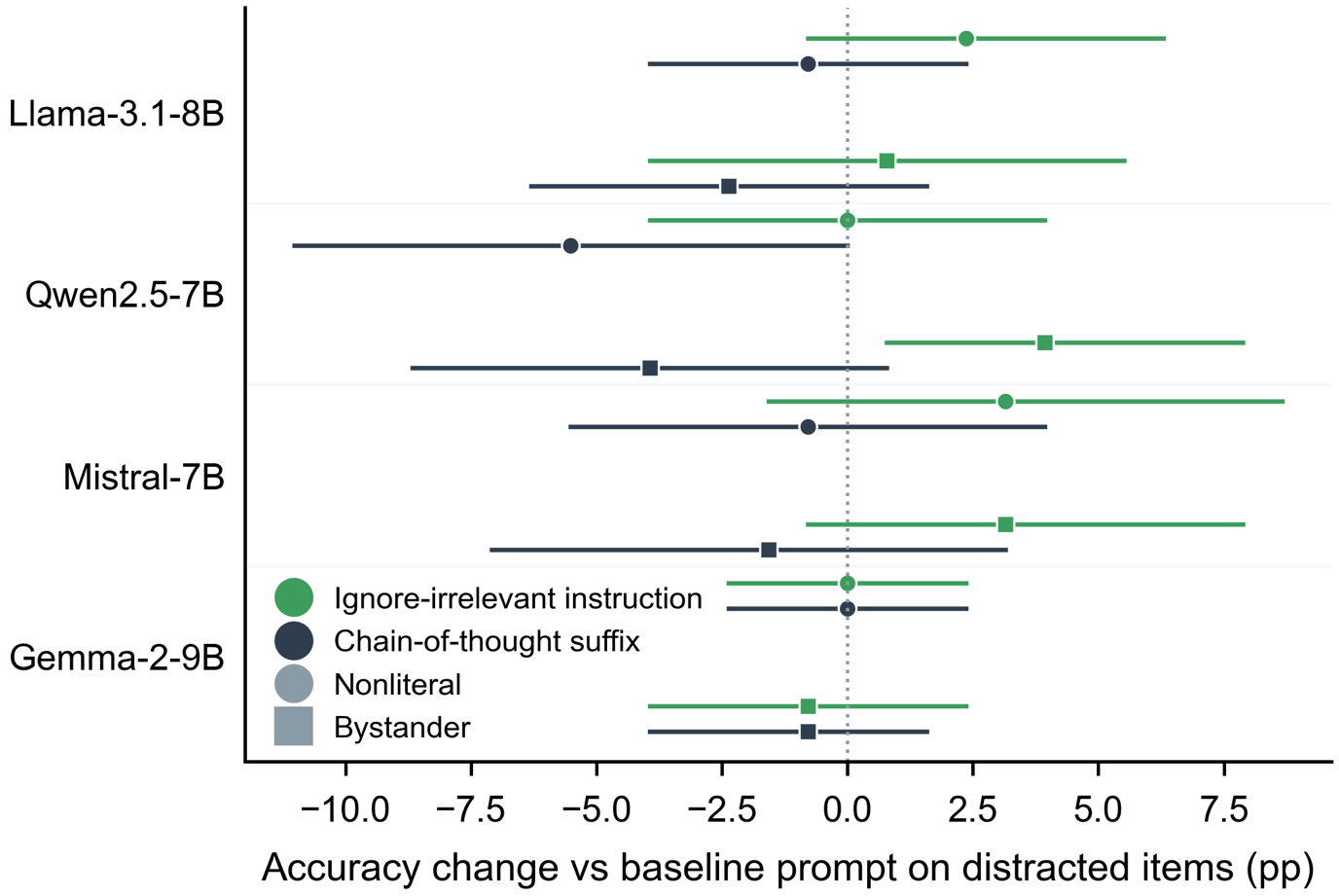}
\par\vspace{6pt}\begingroup\small\raggedright
\textbf{Extended Data Fig. 5 | }Effects of prompting on the mechanistic probe. Accuracy change relative to the baseline prompt on the 127 held-out items with an instruction to ignore irrelevant information or a chain-of-thought preamble, for each model and perturbation family; n = 127 items; error bars are 95\% paired bootstrap CIs.\par\endgroup
\end{figure}
\begin{figure}[!htbp]
\centering
\includegraphics[width=\textwidth,height=.37\textheight,keepaspectratio]{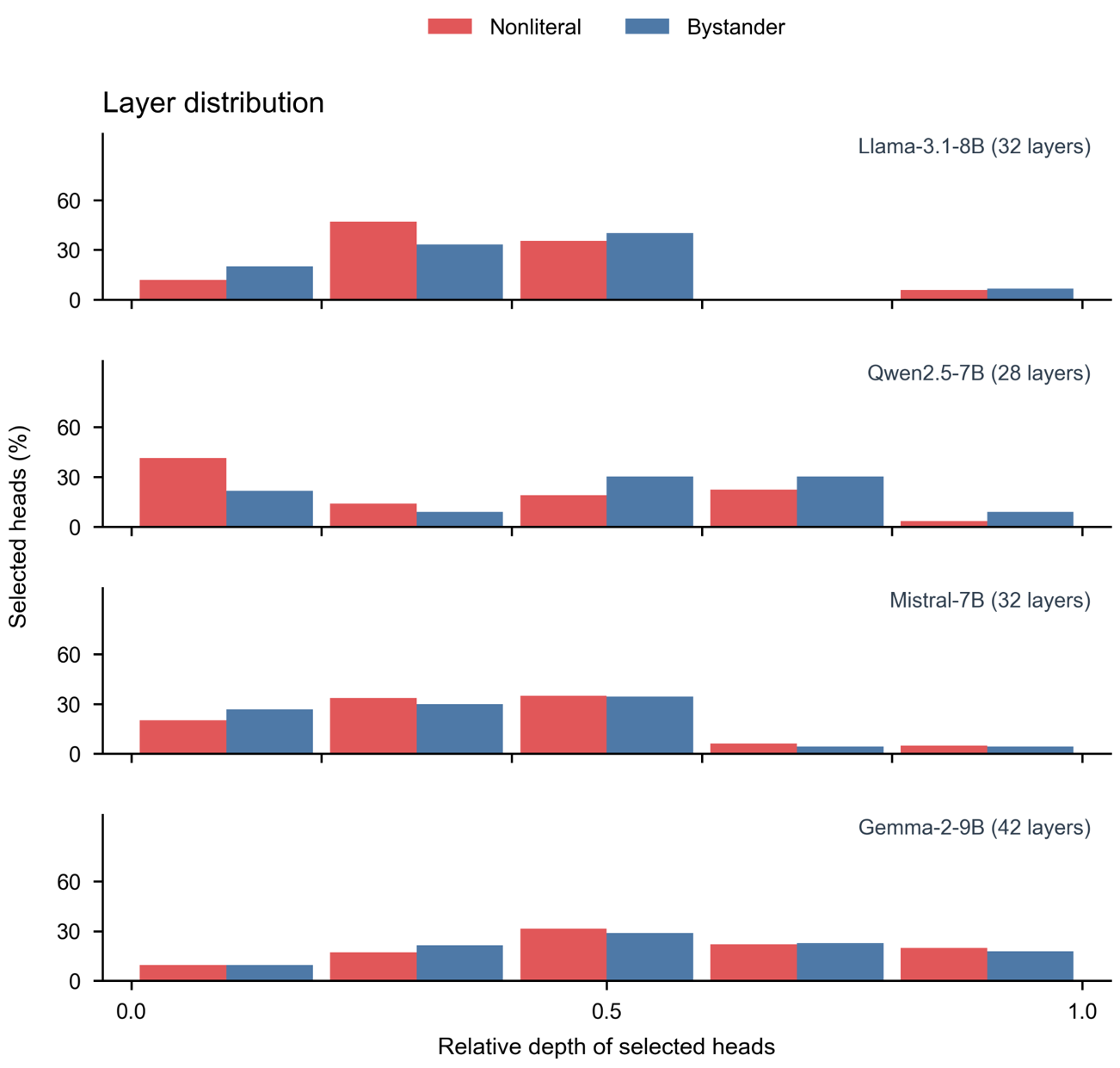}
\par\vspace{6pt}\begingroup\small\raggedright
\textbf{Extended Data Fig. 6 | }Depth distribution of selected attention heads. Histograms show the fraction of selected heads in five equal bins of relative model depth for nonliteral (red) and bystander (blue) perturbations. Relative depth is normalized within each model; model names and total layer counts are shown above each panel.\par\endgroup
\end{figure}
\begin{figure}[!htbp]
\centering
\includegraphics[width=\textwidth,height=.37\textheight,keepaspectratio]{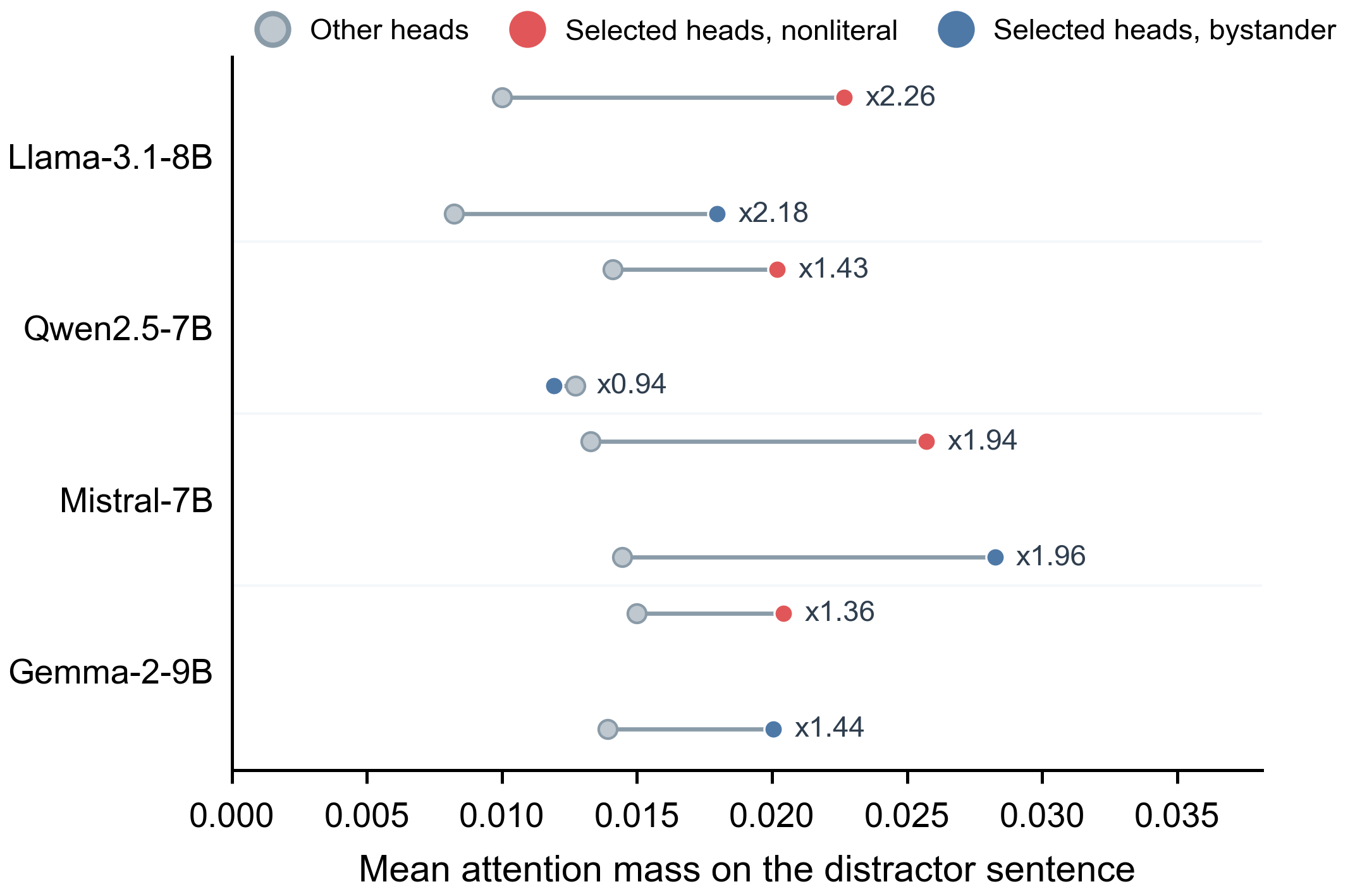}
\par\vspace{6pt}\begingroup\small\raggedright
\textbf{Extended Data Fig. 7 | }Attention to incidental information. Mean attention mass on the inserted sentence in selected versus other heads, measured from the answer-prediction position on the 127 held-out items for each model and perturbation family. Grey markers show other heads; red and blue markers show selected heads for nonliteral and bystander insertions, respectively. Labels give the selected-to-other attention ratio. This diagnostic is descriptive; no confidence intervals are shown.\par\endgroup
\end{figure}
\FloatBarrier\Needspace{10\baselineskip}\noindent{\small \textbf{Extended Data Table 1 }| Interference heads selected by contrastive causal head gating. Heads selected per model and perturbation family, total heads and the Jaccard overlap between the two families' head sets.\par}\nopagebreak\vspace{2pt}
\begingroup\small\setlength{\tabcolsep}{4pt}\renewcommand{\arraystretch}{1.12}
\begin{longtable}{@{}>{\raggedright\arraybackslash}p{0.23\dimexpr\textwidth-8\tabcolsep\relax}>{\raggedright\arraybackslash}p{0.15\dimexpr\textwidth-8\tabcolsep\relax}>{\raggedright\arraybackslash}p{0.22\dimexpr\textwidth-8\tabcolsep\relax}>{\raggedright\arraybackslash}p{0.22\dimexpr\textwidth-8\tabcolsep\relax}>{\raggedright\arraybackslash}p{0.18\dimexpr\textwidth-8\tabcolsep\relax}@{}}
\toprule
\textbf{Model} & \textbf{Total heads} & \textbf{Nonliteral selected} & \textbf{Bystander selected} & \textbf{Jaccard overlap} \\
\midrule
\endfirsthead
\toprule
\textbf{Model} & \textbf{Total heads} & \textbf{Nonliteral selected} & \textbf{Bystander selected} & \textbf{Jaccard overlap} \\
\midrule
\endhead
\bottomrule\endfoot
Llama-3.1-8B & 1024 & 17 (1.7\%) & 15 (1.5\%) & 0.78 \\
\addlinespace[6pt]
Qwen2.5-7B & 784 & 58 (7.4\%) & 56 (7.1\%) & 0.39 \\
\addlinespace[6pt]
Mistral-7B & 1024 & 143 (14.0\%) & 160 (15.6\%) & 0.70 \\
\addlinespace[6pt]
Gemma-2-9B & 672 & 146 (21.7\%) & 159 (23.7\%) & 0.62 \\
\end{longtable}
\endgroup
Supplementary Information. Supplementary Note 1: verbatim prompts (organ-system labelling; exchange generation for both perturbation families; SOAP-note instruction; paired contamination judge; attribution judge; quality judge; log-probability scoring template). Supplementary Note 2: failure-mode classification rules and further verbatim examples. Supplementary Note 3: encounter-level sensitivity analyses and exploratory comparisons.

\FloatBarrier
\section{\textbf{Methods}}
\textbf{Study design. }Every clean input (a transcript, recording or question) was paired with a distracted version containing the same patient information and reference material, differing only by the inserted content. The documentation and mechanistic arms shared two perturbation families. The prespecified primary documentation endpoint was the distracted-minus-clean rate of target-specific contamination in generated notes. Mechanistic endpoints were the fraction of heads selected by CCHG, the accuracy change after patching selected heads with clean activations relative to size-matched random heads, and the change in clean accuracy after suppressing selected heads. Attribution, failure-mode classification, note-quality scores, sensitivity analyses and learned per-head correction were exploratory. No data were collected from human participants. All corpora are public and contain de-identified or mock encounters. No institutional ethics determination was obtained for this analysis of public datasets.

\textbf{MedDistractNotes corpus. }We used 577 patient--clinician dialogues from two public corpora released under CC BY 4.0. From ACI-Bench\textsuperscript{18} we selected encounters from the ``aci'' source in the official validation set (11) and in the three test sets (22 each; 77 in total); training encounters and the other two sources were not used. From MTS-Dialog\textsuperscript{19} we used Test Set 1 (200), Test Set 2 (200) and the validation set (100), whose dialogues are short exchanges corresponding to a single note section; the corresponding section text served as the reference note. ACI-Bench transcripts label speakers in square brackets (``[doctor]'', ``[patient]'') and cover a whole visit (median 6,538 characters); MTS-Dialog transcripts use ``Doctor:'' and ``Patient:'' labels and are shorter (median 350 characters).

\textbf{Organ-system labelling. }GPT-5.4 (temperature 0, JSON output, September 7, 2026) assigned an organ-system label to each of the 577 encounters using the clean transcript and reference note. It selected from twelve systems (musculoskeletal, cardiovascular, respiratory, gastrointestinal, neurological, genitourinary, endocrine, dermatological, psychiatric, ear--eye--dental, haematological--oncological, obstetric--gynaecological) or a ``general'' category for excerpts with no clinical content, such as an allergy list or a social history. The label was inferred from the chief complaint or any condition, symptom, examination, medication or question in the excerpt. The 177 excerpts labeled ``general'' were subsequently randomly assigned an organ system using a fixed seed. Labels and rationales will be released with the benchmark.

\textbf{Insertion of incidental exchanges. }For each encounter and each perturbation family, we asked GPT-5.4 (JSON output, temperature 0.3, September 7, 2026) to generate a three- to four-turn small-talk exchange using the transcript's speaker labels, alongside a one-sentence summary and a suggested insertion point. The pipeline assigned the content of each aside from the visit's organ system. Bystander content was a non-communicable condition or elective procedure of a named third party, drawn from a table of 67 conditions across the twelve systems (for example a coworker's coronary stent in a cardiovascular visit, a sister's knee replacement in a musculoskeletal visit, a friend's father's lithotripsy in a genitourinary visit). To reduce the chance of introducing hereditary risk, relatives were only used for injuries and elective procedures; conditions with any heritable component were assigned to non-relatives (e.g., coworkers, neighbours). Nonliteral content used a medical term as a title, name or figure of speech. We selected from 24 items mapped to relevant organ systems (e.g., the band ``The Strokes'' in a neurological or cardiovascular visit, a racehorse named ``Palpitation'', a coffee shop called ``Delirium'', a sourdough starter ``in remission'', a rowing boat named ``Cataract'').

The prompt required the patient, or an accompanying family member or colleague, to raise the topic. The clinician was to reply conversationally without giving advice or announcing a return to the visit. The exchange could include one treatment detail about the other person (a procedure, device or medication name), but no symptom, condition, exposure or concern of the patient's own. A fixed list of words that would suggest contagion, exposure or heredity was forbidden. Automated checks assessed turn count, speaker labels, length and inclusion of the assigned content. They also checked for forbidden words, patient statements about their own symptoms or body parts, and clinician-initiated openings. Exchanges that failed were regenerated (mean 1.02 and 1.15 attempts per bystander and nonliteral aside). One MTS-Dialog excerpt consisting of a single clinician sentence with no second speaker was excluded.

The final corpus contained 576 bystander and 576 nonliteral exchanges, spanning 66 conditions and 24 items. Sciatica was the most frequent bystander condition (21 of 576 exchanges, 3.6\%); a football team's ``hernia'' of a defence was the most frequent nonliteral item (50 of 576, 8.7\%). Bystander and nonliteral exchanges averaged 3.3 and 3.6 turns and 265 and 255 characters. Within MTS-Dialog, insertions accounted for a median 43\% of the distracted transcript. The exchange was spliced after the model-chosen turn in the transcript's own speaker-label format (``[doctor]''/``[patient]'' in ACI-Bench, ``Doctor:''/``Patient:'' in MTS-Dialog), so that the clean and distracted transcripts differed only by the inserted lines; the frozen exchanges, the organ-system labels and both generator prompts will be released with MedDistractNotes (Supplementary Note 1).

\textbf{Note generation. }Llama-3.1-8B-Instruct (revision 0e9e39f), Qwen2.5-7B-Instruct (a09a354), Mistral-7B-Instruct-v0.3 (c170c70) and Gemma-2-9B-it (11c9b30) generated one SOAP progress note for every clean and distracted transcript in MedDistractNotes (bfloat16 weights, greedy decoding, at most 1,024 new tokens, batches of two with left padding; PyTorch 2.5.1, transformers 4.57.6, one A100 80 GB GPU per job). We prompted each model to write a concise SOAP note based solely on the transcript. The instruction and transcript formed a single user turn in the model's native chat template, which supplied the beginning-of-sequence token where defined (Gemma-2, Llama and Mistral). The factorial design of 576 encounters, two perturbation families and four models yielded 4,608 clean--distracted note pairs.

\textbf{Frontier models. }GPT-5.4, GPT-5.6 Sol, Claude Opus 5 and Claude Fable 5.1 (OpenAI and Anthropic APIs, September 2026; through the providers' batch endpoints) generated notes from the same clean and distracted transcripts. Each request used the same instruction and transcript in a single user message, with the provider's default reasoning or thinking setting. We set temperature to 0 where supported; Claude 5 series models do not accept this setting. Output caps were 1,024 tokens for GPT models and 4,096 for Claude models. Two saved note records were empty, one from a failed GPT-5.4 request and one from a GPT-5.6 Sol response that exhausted its token cap during reasoning. Both remained in the primary analysis; excluding their pairs did not change the rounded frontier incorporation rate (Supplementary Note 3).

\textbf{Contamination judging. }Claude Sonnet 5 evaluated contamination using a paired protocol (Anthropic API; September 2026). For each pair, it received the clean transcript, a one-sentence summary of the inserted content as the evaluation target, and the two notes presented in a randomized order. The instructions specified that the target was an evaluation criterion, that each note should be scored independently, that paraphrases and clinical consequences counted, and that unrelated hallucinations did not. For each note, the judge returned a binary judgement of clinically meaningful incorporation and a severity score from 0 to 3. Clean notes served as false-positive controls within encounters. 

GPT-5.4 (temperature 0) independently scored each note using the transcript and reference note for clinical correctness, completeness and succinctness on five-point scales, alongside a binary hallucination flag and overall quality score.

To test whether showing both notes affected scores, we used the same Claude Sonnet 5 judge to score all 18,432 saved notes one at a time. Each request contained the original clean transcript, the insertion summary and one note, without its condition label. The incorporation criterion and generation settings matched the paired protocol. We compared flag rates and agreement between protocols, with 95\% intervals from 5,000 whole-encounter bootstrap resamples that retained all models and perturbations together (Extended Data Fig. 3). Full prompts, run settings and results are provided in Supplementary Notes 1 and 3.

\textbf{Attribution. }For each distracted note flagged for incorporation, a second Claude Sonnet 5 call received the clean transcript, the description of the inserted content and the note. It classified the note as presenting the content as clinical information about the patient, recording it with the correct meaning (e.g., as another person's information or a non-clinical remark), or not mentioning it. It also assessed whether any assessment, diagnosis or plan item for the patient rested on the content. Empty replies were retried once. We report misattribution, correctly attributed clinical use and recorded-only mention as percentages of distracted notes in each cohort. The prompt appears in Supplementary Note 1. Of the 1,612 frontier notes flagged for incorporation, the attribution judge marked 15 as absent and five remained unresolved. The corresponding counts among 2,309 open-weight notes were 17 and ten. These notes remain in the denominator but outside the three attribution categories.

\textbf{Failure-mode analysis. }For each distracted note in MedDistractNotes, we matched the aside's content words (condition, procedure, device, medication or figurative item) to note lines. Each matching line was assigned to the nearest preceding section header (chief complaint or history of present illness; past medical, surgical or medication history and review of systems; family or social history; other subjective; objective; assessment; plan), recognizing the ``S/O/A/P'', ``Subjective/Objective/Assessment/Plan'' and long-form header styles the models use. This procedure located content in 94--99\% of notes incorporating bystander exchanges and 54--80\% of notes incorporating nonliteral exchanges, which were often paraphrased.

Misattributed and acted-on notes were then grouped into failure modes by a fixed, ordered list of keyword rules applied to the attribution judge's rationale, the contamination judge's rationale and the matching note lines: small talk documented as a mental-status or psychiatric sign (tangential, off-topic, cognitive, mental status, confusion, anxiety, panic, psychiatric); another person's condition becoming the patient's complaint or diagnosis (chief complaint, primary diagnosis, the patient's own condition, surgical history, status post); recorded as family history or a risk factor (family history, hereditary, familial, risk factor); read as an unvoiced concern with screening or counselling added (unvoiced, screen, counsel, inquire, interest in, explore); recorded as the patient's own activity or social history (activity, social history, recreational, plays, participates); and a figurative remark recorded as a fact about the patient (converts, recasts, misrepresents, literal, factual, fabricated). The first matching rule determined the mode; unmatched notes were classified as other. The rules, the per-note assignments and verbatim examples for every mode and model will be released (Supplementary Note 2), and the examples in Table 1 are reproduced from the notes without editing beyond abridgement.

\textbf{Statistics for MedDistractNotes. }Each clean--distracted note pair contributed a paired contamination difference. The reported bootstrap resampled note pairs; pooled analyses did not cluster all models and perturbations from the same encounter. The primary estimand is the pooled (encounter-weighted) mean difference over all note pairs of a cohort (4,608 per cohort) and separately by model, perturbation family, corpus and their combinations, with 95\% percentile bootstrap confidence intervals from 2,000 note-pair resamples\textsuperscript{28}. An additional analysis resampled whole encounters, retaining all models and perturbations together (5,000 resamples; Supplementary Note 3). The unweighted mean over the 40 model \ensuremath{\times} dataset-split \ensuremath{\times} perturbation strata is reported alongside because the strata differ in size from 11 to 200 encounters. Attribution outcomes and failure modes are reported as percentages of distracted notes in each cohort with the same note-pair bootstrap. We analysed quality scores as paired differences using the same procedure. The association between transcript length and contamination was assessed within MTS-Dialog by Spearman correlation and by pooled differences within quintiles of clean-transcript length, and across the cohort by a logistic regression of distracted-note contamination on the relative position of the inserted exchange (turn index divided by the number of turns) and log10 transcript length. Agreement between paired and single-note protocols is reported as percentage agreement and Cohen's \ensuremath{\kappa}.

\textbf{MedDistractAudio benchmark. }We tested acoustic contamination using PriMock57, a public corpus of 57 mock primary-care consultations recorded as separate 16-kHz doctor and patient channels with utterance-level transcripts and clinician notes (CC BY 4.0)\textsuperscript{21}. For each consultation A, we combined the two channels into one track and selected two donor consultations B with a fixed seed, excluding A. From each donor, we selected a 5--15 s window with the densest clinical content, measured by keyword count in the human transcript. We overlaid this segment at a fixed onset within the middle 60\% of A, at \ensuremath{-}20, \ensuremath{-}15, \ensuremath{-}10 and \ensuremath{-}5 dB relative to A's root-mean-square level. Mixes were peak-normalized. The unmodified track served as the clean control, yielding 57 clean and 456 mixed recordings.

Clean and mixed audio were transcribed with Whisper large-v3\textsuperscript{22} (greedy decoding, English, 30-s chunks, no diarization), and the same four open-weight models generated SOAP notes from the transcripts with the protocol above. Transcription leakage was defined as the fraction of content words of the overlaid segment (alphabetic, at least four characters, stop words removed) that appeared in the mixed transcript but not in the clean transcript of A. Contamination was scored with the paired judge described above, with the human transcript of the overlaid segment as the target, prefixed as content from a different patient's consultation. Prespecified endpoints were the pooled contamination difference at \ensuremath{-}10 dB (primary), transcription leakage and contamination by level, and contamination conditional on leakage.

The consultation was the resampling unit, with its donors and models resampled together. We estimated mixed-minus-clean paired differences with 95\% percentile bootstrap confidence intervals over consultations (2,000 resamples), and summarized leakage at each level using the same bootstrap. We assessed the association between leakage and contamination by Spearman correlation across mixes. Two of 1,824 note pairs were excluded because judge outputs could not be parsed.

\textbf{MedDistractQA items used as the mechanistic probe. }The mechanistic experiments used the items of MedDistractQA, our released benchmark\textsuperscript{7}, which extends the 1,273-item test split of MedQA-USMLE (four-option version)\textsuperscript{23}. For each item, one nonliteral and one bystander sentence were generated with GPT-4o (temperature 0.6, February 2025) from a prompt with five fixed example sentences, the clean question and a ``clinical topic'' defined deterministically as the text of the first incorrect answer option in letter order, to make the inserted sentence reuse vocabulary from an incorrect option; the sentence was inserted immediately before the final sentence of the question stem and conveys no information about the patient. The fixed examples led the generated sentences to reuse a small set of frames. Most bystander sentences follow the form ``the patient's aunt mentioned that her friend's parrot \ldots{}''. Each clean--distracted pair nevertheless contains the same options and correct answer. The dataset is released with item identifiers, clean questions, both distracted variants and the inserted sentences.

\textbf{Mechanistic evaluation. }The four note-generating models were evaluated on MedDistractQA by conditional log-probability. Each item was formatted as a fixed instruction, the question, the four lettered options and ``Answer:'', wrapped in the model's native chat template (question and options as the user turn, which supplies the beginning-of-sequence token where the tokenizer defines one) with ``Answer:'' opening the assistant turn; the model's prediction is the option letter with the highest mean log-probability of its single-token target. Items were split deterministically (seed 0) into 1,146 training items, used only to fit gates, and 127 held-out items. All intervention results in the main text were evaluated on the 127 held-out items for each of the four models. Runs used bfloat16 weights on one A100 80 GB GPU.

\textbf{Contrastive causal head gating. }CCHG extends causal head gating\textsuperscript{24} to a paired objective. A multiplicative gate \textit{g}\textsubscript{h} = \ensuremath{\sigma}(\textit{z}\textsubscript{h}) is applied to the output of every attention head before the output projection. Gate logits are initialized at zero (gates at 0.5), and the loss on a batch of paired items is the mean over items of [log p(\textit{y}\textsubscript{i} | \textit{x}\textsubscript{i}\textsuperscript{clean}) \ensuremath{-} log p(\textit{y}\textsubscript{i} | \textit{x}\textsubscript{i}\textsuperscript{distracted})]\textsuperscript{29} plus \ensuremath{\lambda} times the mean over heads of (1 \ensuremath{-} \textit{g}\textsubscript{h}), with \ensuremath{\lambda} = 5. The first term reduces the difference in correct-answer log-probability between the paired inputs. The second pulls gates towards one, penalizing attenuation. Gates are applied in both forward passes. The objective can therefore decrease either by raising distracted or by lowering clean log-probability. Selected heads contribute to the difference in answer probability between the paired inputs. We fixed \ensuremath{\lambda} a priori at 5 for all models and families without tuning. The selected fraction depends on \ensuremath{\lambda} and compares models under a common penalty; it is not an intrinsic model property.

Gates were optimized with Adam for 500 updates at a learning rate annealed linearly from 10\textsuperscript{\ensuremath{-}2} to 10\textsuperscript{\ensuremath{-}3}, an effective batch of 16 pairs (micro-batches of two pairs with eight gradient-accumulation steps), gradient-norm clipping at 1.0, logits clamped to \ensuremath{\pm}6 and sequences truncated from the left to 512 tokens, separately for each model and perturbation family, with one training seed. Heads with a final gate below 0.5 are termed interference heads. Overlap between the nonliteral and bystander head sets is the Jaccard similarity, and head depth is the layer index divided by the number of layers minus one.

To measure seed stability, the gates were retrained with optimization seeds 1--3 under identical data, hyperparameters and update count (for Gemma-2-9B, micro-batches of one pair with 16 accumulation steps replaced the 2 \ensuremath{\times} 8 configuration because eager attention exceeded GPU memory; the effective batch and objective are unchanged), and reproducibility was summarized by the mean pairwise Jaccard similarity between seeds, the Jaccard similarity of each seed with the published seed-0 set and the fraction of published heads selected by at least two seeds; the expected Jaccard similarity of two random sets of the observed sizes is given for reference.

\textbf{Activation patching and suppression. }For clean-activation patching, we replaced selected-head outputs on the distracted input with outputs recorded on the clean input. Token positions were aligned using the longest common prefix and suffix of the two sequences. Inserted tokens had no clean counterpart and were left unpatched. Controls patched an equal number of randomly chosen heads (drawn with a fixed seed from the non-selected heads), or all heads. The comparator is the same pipeline's accuracy on the distracted held-out items without intervention.

Suppression multiplies the outputs of the selected heads by a factor of 1, 0.75, 0.5, 0.25 or 0 on both clean and distracted inputs, with 0 corresponding to removal; as a control, a size-matched set of heads drawn at random (fixed seed) from the non-selected heads was scaled in the same way. Changes are computed relative to the factor-1 evaluation of the same pipeline.

For patching and suppression, each intervention was compared with the unmodified evaluation under the same chat-template and scoring protocol.

As a descriptive diagnostic, we also measured the attention from the position at which the answer is predicted to the inserted tokens, in selected versus other heads. This diagnostic used the same 127 held-out items for all four models. An exploratory learned per-head correction (a small network trained to map the selected heads' distracted activations to their clean activations) changed held-out accuracy by at most 0.8 percentage points and is not used to support any claim. Prompting comparisons in these models used the same log-probability scoring with an added instruction to ignore irrelevant information or with a chain-of-thought preamble.

\textbf{Mechanistic statistics. }Accuracy changes on the 127 held-out items are paired differences with 95\% percentile bootstrap confidence intervals over items (2,000 resamples) and exact McNemar tests; Benjamini--Hochberg correction\textsuperscript{30} was applied within the family of eight baseline clean-versus-distracted changes (Fig. 4a), of which seven remained significant; patching and suppression comparisons are reported with uncorrected exact McNemar \textit{P} values. Comparisons across models are descriptive.

\textbf{Reproducibility and code. }MedDistractNotes and MedDistractAudio will be released with all prompts, frozen generated artifacts (exchanges and summaries for both perturbation families, organ-system labels and MedDistractQA sentences), per-item results, generated notes, judge outputs, failure-mode assignments, and analysis and figure code at \url{https://github.com/nyuolab/llm_distract}, with commands to regenerate every table and figure from the released artifacts on a CPU and to rerun each experiment from scratch. MedDistractQA is available at \url{https://huggingface.co/datasets/KrithikV/MedDistractQA}.

\textbf{Reporting summary. }Research-design information is provided in the accompanying reporting summary.

\end{document}